\documentclass[11pt,a4paper]{article}

\usepackage[utf8]{inputenc}
\usepackage[T2A,T1]{fontenc}
\usepackage[russian,english]{babel}
\usepackage{times}
\usepackage[margin=1in]{geometry}
\usepackage{hyperref}
\usepackage{url}
\usepackage{booktabs}
\usepackage{amsfonts}
\usepackage{amsmath}
\usepackage{nicefrac}
\usepackage{microtype}
\usepackage{graphicx}
\usepackage{subcaption}
\usepackage{pgfplots}
\usepackage{tikz}
\usepackage{tabularx}
\usepackage{multirow}
\usepackage{multicol}
\usepackage{listings}
\usepackage{xcolor}
\usepackage{float}
\usepackage{natbib}
\usepackage{titlesec}
\usepackage{fancyhdr}
\usepackage{tcolorbox}
\usepackage{enumitem}
\usepackage{caption}
\usepackage{needspace}
\usepackage{parskip}

\pgfplotsset{compat=1.18}

\definecolor{codegreen}{rgb}{0,0.6,0}
\definecolor{codegray}{rgb}{0.5,0.5,0.5}
\definecolor{codepurple}{rgb}{0.58,0,0.82}
\definecolor{backcolour}{rgb}{0.95,0.95,0.92}
\lstdefinestyle{pythonstyle}{
    language=Python,
    backgroundcolor=\color{backcolour},   
    commentstyle=\color{codegreen},
    keywordstyle=\color{magenta},
    numberstyle=\tiny\color{codegray},
    stringstyle=\color{codepurple},
    basicstyle=\ttfamily\footnotesize,
    breakatwhitespace=false,         
    breaklines=true,                 
    captionpos=b,                    
    keepspaces=true,                 
    numbers=left,                    
    numbersep=5pt,                  
    showspaces=false,                
    showstringspaces=false,
    showtabs=false,                  
    tabsize=2
}
\definecolor{zehnred}{RGB}{163, 55, 55}
\definecolor{zehnredlight}{RGB}{248, 220, 235}
\definecolor{zehnredmid}{RGB}{195, 100, 100}
\definecolor{linkblue}{RGB}{0, 102, 204}

\definecolor{othermodel}{RGB}{232, 218, 228}
\definecolor{othermodelborder}{RGB}{210, 192, 208}

\definecolor{sorofamily}{RGB}{196, 120, 155}
\definecolor{sorofamilyborder}{RGB}{170, 95, 130}

\definecolor{sorofp8}{RGB}{178, 75, 120}
\definecolor{sorofp8border}{RGB}{155, 55, 100}

\definecolor{soromain}{RGB}{155, 30, 95}
\definecolor{soromainborder}{RGB}{130, 15, 75}

\hypersetup{
    colorlinks=true,
    linkcolor=linkblue,
    citecolor=linkblue,
    urlcolor=linkblue,
    bookmarksnumbered=true,
}

\titleformat{\section}{\Large\bfseries}{\thesection}{1em}{}
\titleformat{\subsection}{\large\bfseries}{\thesubsection}{1em}{}
\titleformat{\subsubsection}{\normalsize\bfseries}{\thesubsubsection}{1em}{}

\renewcommand{\headrulewidth}{0pt}
\newtcolorbox{abstractbox}{
    colback=zehnredlight,
    colframe=zehnredlight,
    arc=3pt,
    boxrule=0pt,
    left=15pt,
    right=15pt,
    top=12pt,
    bottom=12pt,
}

\begin{document}
\thispagestyle{empty}
\raggedbottom


\begin{tcolorbox}[
  colback=zehnredlight,
  colframe=zehnredlight,
  arc=3mm,
  boxrule=0pt,
  left=0.7cm,
  right=0.7cm,
  top=0.7cm,
  bottom=0.7cm,
]

\noindent
\includegraphics[height=1.1cm]{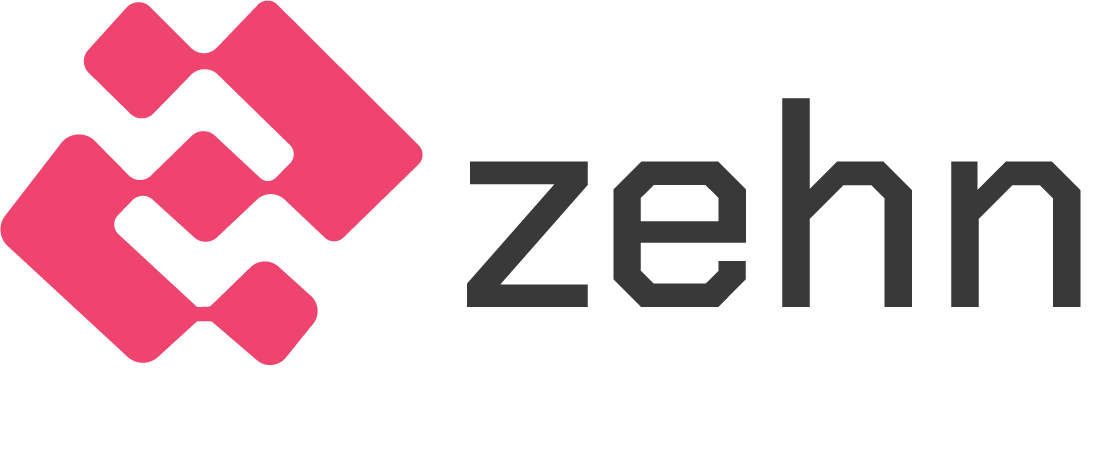}

\vspace{0.8cm}

{\raggedright\fontsize{22}{26}\selectfont\bfseries
Soro: A Lightweight Foundation Model and Chatbot for Tajik\par}

\vspace{0.6cm}

{\raggedright\normalsize Stanislav Liashkov, Haitz Sáez de Ocáriz Borde, Azizjon Azimi, Khushbakht Shoymardonov,\\Shuhratjon Khalilbekov, Bonu Boboeva\par}
\vspace{0.15cm}
{\raggedright\normalsize\textit{zehnlab.ai}\par}

\vspace{0.6cm}

\normalsize
We present Soro, a family of Tajik-specialized conversational large language models (LLMs) designed for real-world deployment under tight compute and connectivity constraints in Tajikistan. Starting from open-weight Gemma~3 checkpoints, we perform Tajik-only continual pretraining on a curated 1.9-billion-token corpus spanning filtered web text, PDF documents, and curriculum-aligned educational materials, followed by supervised instruction tuning on 40K Tajik teacher-style examples. To enable rigorous evaluation despite the limited coverage of Tajik in standard benchmarks, we introduce a suite of Tajik benchmarks covering general knowledge, linguistic competence, and school- and university entrance-exam domains, and we open-source them on Hugging Face. Across these Tajik benchmarks, Soro substantially outperforms same-size Gemma~3 baselines while retaining strong English performance on standard datasets. We further show that FP8 and INT4 quantization of Soro preserves most Tajik-language gains while reducing memory requirements for edge deployment, supporting an ongoing education-sector pilot and planned scale-out across schools in Tajikistan.

\vspace{0.4cm}

{\color{zehnred}\textbf{Date:}} April 2026\\
{\color{zehnred}\textbf{Website:}} \url{https://sorollm.tj}

\end{tcolorbox}

\section{Introduction}

Large language models (LLMs) have achieved strong performance across a wide range of natural language tasks, yet their capabilities remain heavily skewed toward high-resource languages such as English~\citep{openai2024gpt4technicalreport,brown2020languagemodelsfewshotlearners} and Chinese~\citep{Guo_2025,yang2025qwen3technicalreport}. Tajik, an Indo-Iranian language written in Cyrillic script and spoken primarily in Tajikistan and neighboring regions of Uzbekistan and Afghanistan, falls well outside this favored group. Closely related to Persian (Farsi) and Dari, Tajik has a rich literary and cultural heritage, yet suffers from a sparse digital corpus, scarce annotated resources, and minimal representation in mainstream multilingual training data.

The consequences are practical: limited language technology coverage restricts access to modern AI tools in education, administration, and cultural preservation in Tajikistan. To address this, we introduce the Soro model family: efficient, Tajik-specialized LLMs built for low-cost deployment under real hardware constraints. Soro is trained via targeted continual pretraining on a curated Tajik corpus followed by instruction tuning, and is designed to be adopted by under-resourced organizations without sacrificing strong Tajik-language performance or broadly useful general capabilities. The result is a concrete example of how focused adaptation of an open foundation model can produce deployable language technology for a low-resource setting.

Our development process centers on three considerations: data curation, parameter-efficient adaptation, and deployment feasibility.

\begin{itemize}[leftmargin=1.5em]
    \item \textbf{Tajik-centric models.} We present \textit{Soro} (27 billion parameters) and \textit{Soro Lite} (12 billion parameters), Tajik-specialized conversational LLMs that, to the best of our knowledge, are among the first Tajik-centric LLMs developed in Tajikistan. The larger 27B variant is deployed via centralized API infrastructure in Dushanbe (the capital of Tajikistan), while the 12B model is optimized for edge or low-connectivity environments, targeting rural regions with limited internet access.
    \item \textbf{Education-sector deployment.} Alongside the model release, we describe a structured education pilot across 100 schools in Tajikistan as part of \textit{Project Soro}, in which students are taught the fundamentals of artificial intelligence as a dedicated school subject. Section~\ref{sec:deployment} details the pilot, its use cases, and preliminary observations from the initial deployment.
    \item \textbf{Tajik benchmarks.} To address the near-complete absence of Tajik evaluation resources, we introduce Tajik-FactQA and multiple Tajik evaluation sets (Tajik MMLU, TajLib, and others covering high-school-level history and literature) in Section~\ref{sec:evaluation}. These benchmarks assess both linguistic competence and culturally grounded factual knowledge.
\end{itemize}

\section{Related Work}

Soro is positioned at the intersection of multilingual foundation models, low-resource language adaptation, and parameter-efficient training. Next, we review relevant related work.

\paragraph{Multilingual foundation models and data imbalance.}
Modern LLMs train on web-scale corpora~\citep{brown2020languagemodelsfewshotlearners,deepseekai2025deepseekv3technicalreport,openai2024gpt4technicalreport}. These datasets are multilingual but dominated by a few languages (e.g., English, Chinese, Spanish, etc), creating data imbalance. This has been analyzed in the context of multilingual pretraining and evaluation in prior works, including the curse of multilinguality and data-mixture trade-offs in massively multilingual LLMs~\citep{chang2023multilingualitycurselanguagemodeling,blevins2024breakingcursemultilingualitycrosslingual,pfeiffer2022liftingcursemultilingualitypretraining,li2025rethinkingmultilingualcontinualpretraining}. Open multilingual efforts like BLOOM~\citep{workshop2023bloom176bparameteropenaccessmultilingual} and the ROOTS corpus~\citep{Laurenccon2023TheBR} stress transparency and broad coverage, but also expose the scarcity of high-quality web text for many languages. Recent curation pipelines (FineWeb2, FinePDFs)~\citep{penedo2025fineweb2pipelinescale,kydlicek2025finepdfs} and educational subsets like FineWebEDU~\citep{lozhkov2024fineweb-edu} broaden coverage to languages such as Tajik, yet raw inclusion alone does not ensure robustness. Filtering, balancing, and domain-aware curation remain necessary. These constraints motivate language-targeted adaptation: specialize models to one low-resource language via continual pretraining and post-training, rather than relying on multilingual generalization, like the Soro model family.

\paragraph{Language-specialized LLMs for low-resource settings.}
Many results have shown that adapting open-weight foundation models to low-resource languages can deliver large gains at modest cost. Instead of training from scratch, common pipelines use continual (domain and language-adaptive) pretraining on curated monolingual data~\citep{dontstoppretraining2020}, then instruction tuning~\citep{wei2022finetuned} for conversational and task performance. Recent regional and low-resource models built on strong backbones (e.g., LLaMA, Mistral) show that targeted adaptation improves fluency and factual grounding in the target language~\citep{touvron2023LLaMA2openfoundation,nguyen2023vinaLLaMALLaMAbasedvietnamesefoundation,Corra2024TeenyTinyLLaMAOT,jiang2023mistral7b,zolkepli2024largemalaysianlanguagemodel}. The approach is practical: reuse strong pretrained representations, then add language-specific data and alignment in post-training. Soro follows this. We start from a multilingual base (Gemma 3~\citep{gemmateam2025gemma3technicalreport}), do Tajik-only continual pretraining to boost competence, then supervised fine-tuning to produce an instruction-following chatbot for local educational and informational needs. This aligns with \emph{Type II Partial AI Sovereignty}~\citep{saezdeocarizborde2025sovereign}: adapt open-weight models to encode local knowledge while keeping local deployment feasible.

\paragraph{Parameter-efficient adaptation and deployment constraints.}
Compute limits dominate low-resource settings; full fine-tuning is often economically infeasible. Parameter-efficient methods, especially LoRA~\citep{hu2022lora} and variants~\citep{pmlr-v235-zhu24c,dettmers2023qloraefficientfinetuningquantized}, are now standard, cutting training cost and hardware needs. They matter most where large GPU clusters are scarce (e.g., developing countries, small institutions). Soro follows this trend and utilizes LoRA for training.

\paragraph{Model merging.}
The growth of task-specific derivatives from open-weight foundations has made model merging a gradient-free alternative to multi-task learning. For models with a shared ancestor, merging combines capabilities without extra training by directly manipulating weights. Early methods used linear averaging~\citep{wortsman2022modelsoupsaveragingweights}; newer work uses task arithmetic~\citep{ilharco2023editingmodelstaskarithmetic}, adding vectors between fine-tuned and base checkpoints. To reduce interference, methods like TIES~\citep{yadav2023tiesmergingresolvinginterferencemerging}, DARE~\citep{yu2024languagemodelssupermario}, and SCE~\citep{wan2024fusechatknowledgefusionchat} sparsify updates and resolve sign conflicts. Related geometric approaches include SLERP~\citep{Shoemake1985} and multi-model extensions~\citep{goddard-etal-2024-arcees}, which preserve rotational structure to trace lower-loss interpolation paths. For Soro, we merge specialized Tajik checkpoints with Gemma 3. Using \texttt{mergekit}~\citep{goddard-etal-2024-arcees}, we choose simple linear merging because our models are highly similar in architecture and task.

\section{Base Model Selection}

The choice of a foundation model for continual pretraining shapes both the quality of language-specific adaptation and the feasibility of deployment under hardware constraints. Given these dual demands, we established the following selection criteria.

\paragraph{Deployment feasibility.} 
The model must be small enough to run efficiently on hardware commonly available in Tajikistan (ideally, single-GPU configurations accessible to educational institutions, government offices, and local organizations). We therefore consider models in the 8B--30B parameter range.

\paragraph{Prior multilingual performance.} 
The model should demonstrate reasonable baseline performance in Tajik or linguistically related languages, indicating that its pretraining corpus included sufficient coverage to establish foundational language representations. We evaluated Gemma~3~\citep{gemmateam2025gemma3technicalreport}, Qwen~3~\citep{yang2025qwen3technicalreport}, Llama~3.1 and 3.2~\citep{grattafiori2024llama3herdmodels}, Qwen~2.5~\citep{qwen2025qwen25technicalreport}, and GPT-OSS~\citep{openai2025gptoss} on our Tajik benchmarks (exact metrics are reported in Section~\ref{sec:evaluation}; see Figure~\ref{fig:avg-tajik-performance}). Gemma~3 performed best overall, followed by Qwen~3; GPT-OSS-20B scored substantially lower than Gemma~3, retaining only about 55\% of the larger Gemma~3 variant's accuracy on the Tajik benchmark suite.
    
\paragraph{Tokenization efficiency.} 
The tokenizer should encode Tajik text efficiently, minimizing the number of tokens per word (known as the fertility rate, i.e., the ratio of tokens to original word count). High fertility increases computational overhead during training and inference and reduces the effective context window for Tajik text. Tokenization efficiency also serves as a reliable proxy for model performance: lower fertility typically indicates that the language was well-represented in the pretraining corpus, allowing the tokenizer to learn more meaningful subword units.

We evaluate tokenizers on two representative text samples: a subset of FineWeb-Edu for English~\citep{lozhkov2024fineweb-edu} and a subset of our curated Tajik corpus. Table~\ref{tab:dataset_stats} presents the characteristics of these evaluation datasets. The Tajik sample has a longer average word length (5.29 characters) compared to English (3.87 characters), reflecting the morphological complexity and Cyrillic script used in Tajik.

\begin{table}[hbpt!]
\centering
\caption{Dataset statistics for tokenization analysis.}
\begin{tabular}{lrrrrl}
\hline
\textbf{Language} & \textbf{Characters} & \textbf{Words} & \textbf{Avg Length} & \textbf{Samples} & \textbf{Source} \\
\hline
English & 81.0M & 11.1M & 3.87 & 20,000 & FineWEB EDU sample \\
Tajik & 106.1M & 16.2M & 5.29 & 20,000 & Our Tajik corpus sample \\
\hline
\end{tabular}
\label{tab:dataset_stats}
\end{table}

Table~\ref{tab:fertility_comparison} compares English and Tajik fertility rates across several foundation models. All models exhibit higher fertility for Tajik, reflecting the general underrepresentation of non-Latin scripts, but the disparity varies considerably. Gemma~3 achieves the most efficient Tajik tokenization, with a substantially lower fertility rate than Llama~3.1, Qwen, and DeepSeek~\citep{dai2024deepseekmoeultimateexpertspecialization} (DeepSeek is evaluated only in this section and produces the worst fertility rates). This efficiency directly expands the effective context window, enabling processing of substantially longer Tajik sequences. Llama models are the most efficient for English but show pronounced degradation on Tajik; Gemma~3 maintains competitive English tokenization while providing superior Tajik representation, making it the most suitable foundation for applications where Tajik coverage is the priority.

\begin{table}[hbpt!]
\centering
\caption{Fertility rates comparison across models and languages.}
\begin{tabular}{lccc}
\hline
\textbf{Model} & \textbf{Vocab Size} & \textbf{Fertility (English)} & \textbf{Fertility (Tajik)} \\
\hline
Llama 3.1-8B & 128,000 & \textbf{1.634} & 2.798 \\
Llama 3.2-3B-Instruct & 128,000 & \textbf{1.634} & 2.798 \\
\textit{\textbf{Gemma 3-12B}} & \textit{\textbf{262,144}} & \textit{{1.776}} & \textit{\textbf{2.380}} \\
Qwen 3-8B & 151,643 & 1.753 & 2.860 \\
Qwen 2.5-14B-Instruct-1M & 151,643 & 1.753 & 2.860 \\
Qwen 3-30B-A3B-Instruct-2507 & 151,643 & 1.753 & 2.860 \\
Ministral 3-8B-Instruct-2512 & 131,072 & 1.742 & 2.696 \\
Deepseek MoE-16B-chat & 100,000 & 1.826 & 2.942 \\
\hline
\end{tabular}
\label{tab:fertility_comparison}
\end{table}

We therefore selected the 12B and 27B Gemma~3 variants~\citep{gemmateam2025gemma3technicalreport} as our foundation models.

\section{Data Curation}
\label{sec:data-curation}

With Gemma~3 established as the base, our next challenge was data. We follow a two-stage training approach: continual pretraining on the base Gemma~3 model~\citep{gemmateam2025gemma3technicalreport}, followed by instruction tuning on conversational data. Data collection proved to be the primary bottleneck in developing Soro. High-quality Tajik data is scarce online, particularly question--answer pairs and conversational dialogues suitable for chatbot training. We therefore combine data from multiple sources, both real and synthetic, as described below.

\subsection{Dataset for Continual Pretraining}
\label{subsec:dataset-for-continual-pretraining}

For continual pretraining, we use a Tajik-only dataset totaling 1.9 billion tokens, drawn from curated Tajik web pages, documents, educational resources, legal documents, books, and news articles, supplemented with high-quality English educational content translated into Tajik.

Our web-based foundation draws from two complementary sources. First, we include the Tajik subset of FineWeb2~\citep{penedo2025fineweb2pipelinescale}, which employs an adaptive processing pipeline that adjusts filtering and deduplication based on language-specific statistics. This matters for Tajik: standard English-centric heuristics would discard many high-quality tokens, and the pipeline also helps distinguish Tajik from closely related Cyrillic-script languages. Second, to provide grounding in scientific and analytical knowledge, we include a translated version of FineWeb-Edu~\citep{lozhkov2024fineweb-edu}. Translating this pedagogically reviewed English subset into Tajik via Gemini~2.5~Flash~\citep{comanici2025gemini25} gives the model coverage of core subjects, including physics, world history, mathematics and social sciences, directly supporting its intended role in the education sector.

Beyond standard web sources, we incorporate the Tajik component of FinePDFs~\citep{kydlicek2025finepdfs}, which captures formal documents, institutional publications, technical reports, and academic papers. These content types are underrepresented in HTML-based web data but are important for educational and institutional use.

The most distinctive component of the corpus is a specialized subset of national educational resources. This material consists of secondary-school learning materials spanning grades 5--11: textbooks, student manuals, and lecture notes, alongside native Tajik literature and translated fiction. Every document was derived from physical school textbooks published by the Ministry of Education and Science, processed via manual transcription or high-fidelity OCR using Tesseract~\citep{smith2007tesseract}. This subset provides the cultural, linguistic, and factual grounding in the Tajik national context that is largely absent from broader web-based corpora. A structured overview of the Tajikistan national secondary-school curriculum for grades 5--11 is provided in Appendix~\ref{app:curriculum}, which contextualizes the subject coverage of this educational subset.

\subsection{Dataset for Instruction Tuning}
\label{sec:data-sft}

Building on the continually pretrained base model (described in more detail in Section~\ref{sec:fine-tuning}), we apply instruction tuning~\citep{wei2022finetuned}. The instruction dataset comprises 40,000 synthetic examples in Tajik covering cultural topics and core secondary-school subjects such as mathematics, science, and history. The examples follow a ``teacher-like'' style: responses are informative and detailed, use accessible language and metaphors, avoid overly formal phrasing, and often include follow-up questions to reinforce student understanding. We generate them with Gemini~2.5~Flash~\citep{comanici2025gemini25}, conditioned on chunks from Tajik secondary-school textbooks (of the same type described in Section~\ref{subsec:dataset-for-continual-pretraining}), Wikipedia, regional vocabulary references, information about historical figures and Tajik history, and multiple-choice question--answer examples.

\paragraph{Quality assurance.} 
To validate the synthetic generation pipeline, we conducted a human quality audit on a stratified random sample of several hundred examples drawn proportionally across subject categories. Native Tajik-speaking reviewers assessed each example along four dimensions: (i)~factual correctness, (ii)~grammatical and lexical naturalness in Tajik, (iii)~stylistic consistency with the target pedagogical tone, and (iv)~absence of hallucinations or source-language interference from the conditioning material. Deficient examples were either manually corrected or removed. The audit revealed a low defect rate, supporting the use of Gemini~2.5~Flash as a reliable generator of Tajik instructional content when appropriately conditioned on authentic source material.

\section{Fine-tuning for the Tajik Language}
\label{sec:fine-tuning}

With the training corpora in place, we now describe how they are used. The pipeline consists of three stages: continual pretraining on the curated Tajik corpus, supervised instruction tuning on the synthetic conversational dataset, and linear merging with the original Gemma~3 checkpoint to recover any general-purpose capabilities lost during adaptation.

\subsection{Continual Pretraining}
\label{sec:cpt}

We perform continual pretraining on Gemma~3-12B and Gemma~3-27B using the corpus described in Section~\ref{sec:data-curation}, producing intermediate checkpoints referred to as Soro Lite Base and Soro Base. The names Soro Lite and Soro are reserved for the final chatbot models after instruction tuning and merging.

\paragraph{LoRA configuration.}
Given computational constraints, we employ LoRA~\citep{hu2022lora} for parameter-efficient fine-tuning. LoRA hypothesizes that weight updates $\Delta W$ during adaptation possess a low intrinsic rank, keeping the pretrained weight matrix $W_0 \in \mathbb{R}^{d \times k}$ frozen while training two low-rank matrices $A \in \mathbb{R}^{r \times k}$ and $B \in \mathbb{R}^{d \times r}$, so that $\Delta W = BA$ with rank $r \ll \min(d, k)$. Restricting updates to a low-rank subspace also acts as an implicit regularizer, limiting trainable parameters and reducing the risk of overfitting, which matters when the available training data are orders of magnitude smaller than typical pretraining corpora.

We apply LoRA to seven target modules within each transformer block: the query, key, value, and output projection matrices (\texttt{q\_proj}, \texttt{k\_proj}, \texttt{v\_proj}, \texttt{o\_proj}) in the self-attention mechanism, and the gate, up, and down projection matrices (\texttt{gate\_proj}, \texttt{up\_proj}, \texttt{down\_proj}) in the feed-forward network. We also fully train the embedding layer (\texttt{embed\_tokens}) to allow the model to refine its token representations for Tajik-specific subword patterns. This LoRA configuration is shared across both training stages.

\begin{table}[hbpt!]
\centering
\caption{Hyperparameters and configuration for continual pretraining.}
\scalebox{0.8}{
\begin{tabular}{ccc}
\hline
\textbf{Model} & \textbf{Soro Lite Base (12B)} & \textbf{Soro Base (27B)} \\
\hline
Starting model & Gemma 3-12B & Gemma 3-27B \\
Max sequence length & 2048 tokens & 2048 tokens \\
Learning rate & 7e-5 & 5e-5 \\
Learning rate scheduler & Cosine & Cosine \\
Training steps & 11,000 & 10,000 \\
Training epochs & $\sim$1 & $\sim$0.89 \\
Warmup steps & 800 & 700 \\
Global batch size & 128 & 128 \\
Batch size per GPU & 2 & 1 \\
Gradient accumulation steps & 8 & 16 \\
LoRA rank & 128 & 64 \\
LoRA alpha & 128 & 64 \\
LoRA dropout & 0.1 & 0.1 \\
Trainable parameters \% & 11.3\% & 6.4\% \\
Precision & BF16 & BF16 \\
Hardware & 8×NVIDIA H200 & 8×NVIDIA H200 \\
Training duration & 1d 13h 52m & 1d 23h 30m \\
\hline
\end{tabular}}
\label{tab:pretrain-hyperparams}
\end{table}

\paragraph{Hyperparameter rationale.}
Table~\ref{tab:pretrain-hyperparams} summarizes the training configurations for both model sizes. We use relatively high LoRA ranks (128 for 12B, 64 for 27B), because continual pretraining requires more substantial weight modification than instruction tuning: the model must acquire new linguistic knowledge rather than merely adjust behavioral patterns. The difference in rank between the two sizes reflects the need to balance trainable parameters against the fixed corpus size. The 27B model has more total parameters, so a lower LoRA rank is sufficient to achieve a comparable trainable-to-total-parameter ratio while also reducing overfitting risk, given that both models are trained on the same 1.9-billion-token corpus. Learning rates ($7\times 10^{-5}$ for 12B, $5\times10^{-5}$ for 27B) were chosen to ensure stable convergence, with the larger model using a lower rate following standard practice, under a cosine schedule. Approximately 7\% of training steps are used for warmup. 

\begin{figure}[H]
    \centering
    \begin{subfigure}[b]{0.48\textwidth}
        \centering
        \includegraphics[width=\textwidth]{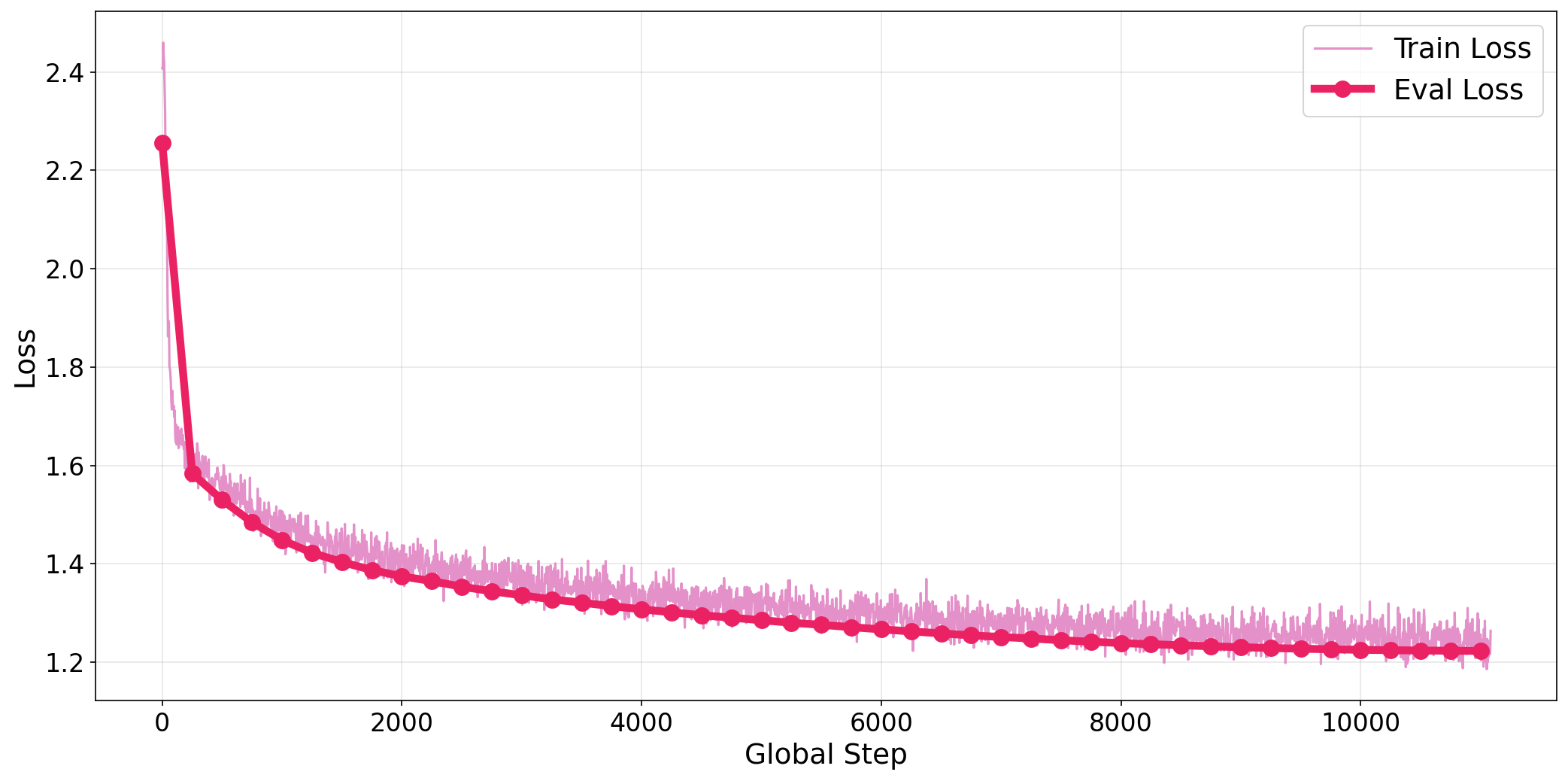}
        \caption{Soro Lite Base (12B)}
        \label{fig:loss-plot-12b}
    \end{subfigure}
    \hfill
    \begin{subfigure}[b]{0.48\textwidth}
        \centering
        \includegraphics[width=\textwidth]{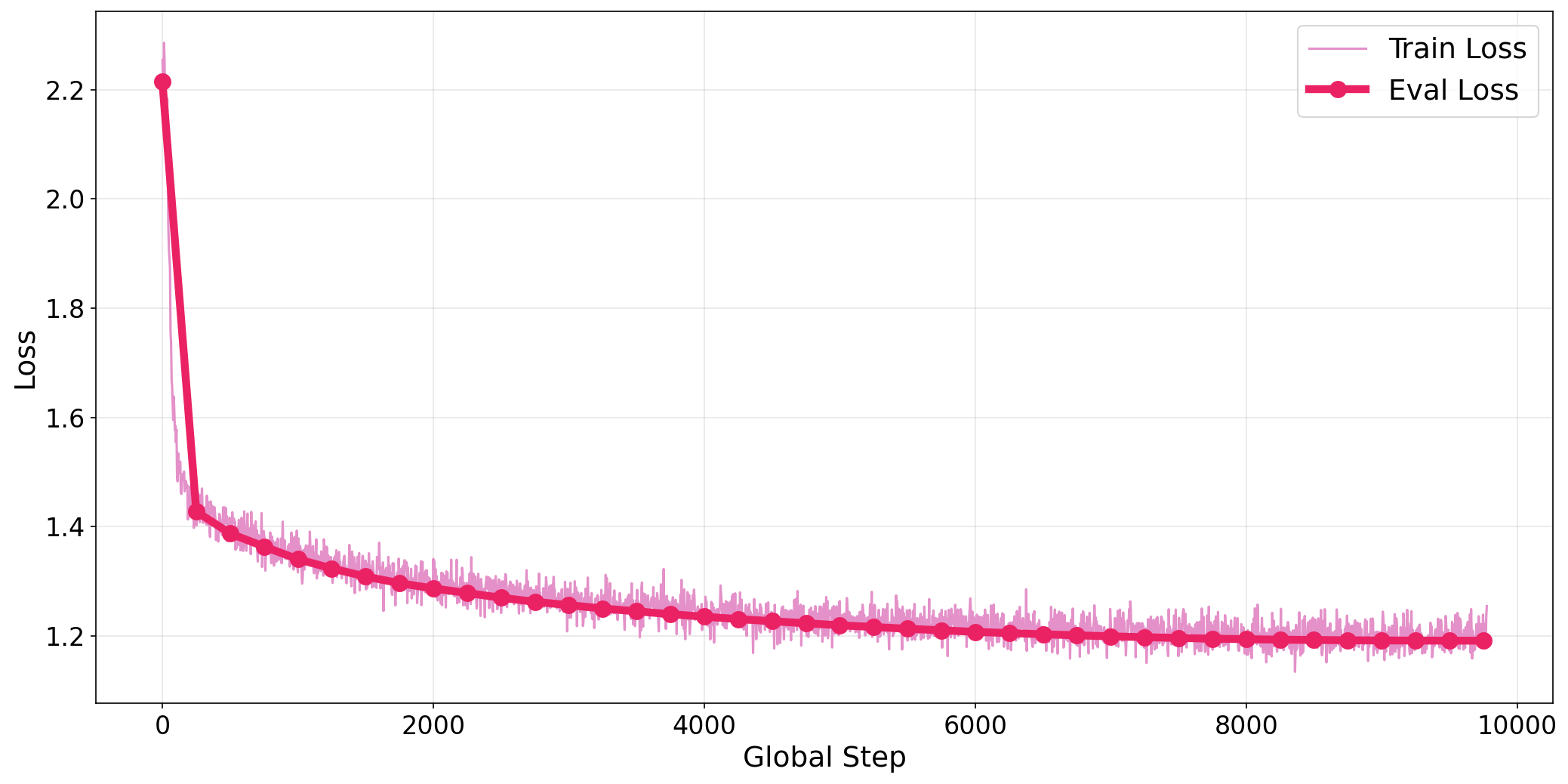}
        \caption{Soro Base (27B)}
        \label{fig:loss-plot-27b}
    \end{subfigure}
    \caption{Training and evaluation loss during continual pretraining of Soro Lite Base and Soro Base (based on Gemma~3 models).}
    \label{fig:combined-loss-plots}
\end{figure}

\paragraph{Convergence.}
Figures~\ref{fig:loss-plot-12b} and~\ref{fig:loss-plot-27b} show the training and evaluation loss curves during continual pretraining for both models. Both models converge smoothly with no sign of overfitting.

\subsection{Instruction Tuning} 
\label{sec:sft}

We perform supervised instruction tuning on both Soro Lite Base and Soro Base using the instruction dataset described in Section~\ref{sec:data-sft}, adapting the continually pretrained checkpoints into conversational assistants capable of instruction following in Tajik.

\begin{table}[hbpt!]
\centering
\caption{Hyperparameters and configuration for instruction tuning.}
\scalebox{0.7}{
\begin{tabular}{ccc}
\hline
\textbf{Setting} & \textbf{Soro Lite} & \textbf{Soro} \\
\hline
Starting model & Soro Lite Base & Soro Base \\
Sequence length & 2048 tokens & 2048 tokens \\
Learning rate & 3e-5 & 1e-5 \\
Learning rate scheduler & Cosine & Cosine \\
Training steps & 10,000 & 9,000 \\
Training epochs & $\sim$2 & $\sim$1.8 \\
Warmup steps & 700 & 700 \\
Global batch size & 8 & 8 \\
Batch size per GPU & 2 & 2 \\
Gradient accumulation steps & 4 & 4 \\
LoRA rank & 32 & 32 \\
LoRA alpha & 32 & 32 \\
LoRA dropout & 0.1 & 0.1 \\
Trainable parameters \% & 8.4\% & 5.62\% \\
Precision & BF16 & BF16 \\
Hardware & 1$\times$NVIDIA H200 & 1$\times$NVIDIA H200 \\
Training duration & 13h 45m 41s & 15h 56m 36s \\
\hline
\end{tabular}}
\label{tab:sft-hyperparams}
\end{table}

\paragraph{Configuration rationale.}
Table~\ref{tab:sft-hyperparams} summarizes the instruction tuning configuration. Several differences from continual pretraining reflect the distinct nature of this task. First, the LoRA rank is reduced from 128/64 to 32 for both model sizes, for two reasons: instruction tuning shapes output behavior rather than acquiring new factual or linguistic knowledge, which requires fewer trainable parameters; and the instruction dataset (40K examples) is far smaller than the pretraining corpus (1.9B tokens), making a lower rank necessary to prevent overfitting. Second, learning rates are reduced ($3\times10^{-5}$ for 12B, $1\times10^{-5}$ for 27B) to prevent catastrophic forgetting of the Tajik knowledge acquired during pretraining. Third, the global batch size is reduced from 128 to 8. We train for approximately two epochs, since behavioral alignment on a 40K-example dataset benefits from repeated exposure to reinforce the desired response style and instruction-following patterns.

\begin{figure}[H]
    \centering
    \begin{subfigure}[b]{0.48\textwidth}
        \centering
        \includegraphics[width=\textwidth]{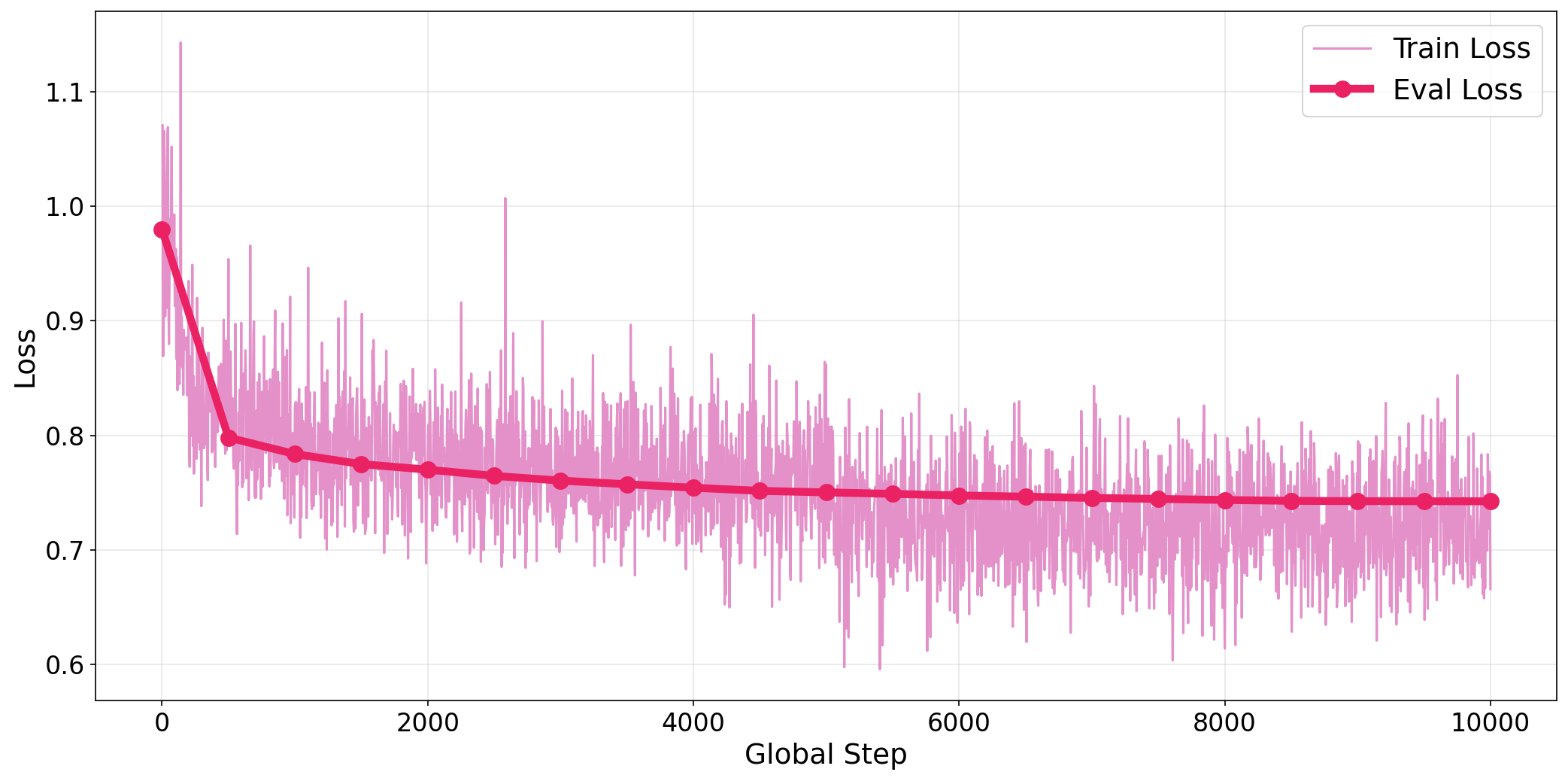}
        \caption{Soro Lite pre-merge 12B}
        \label{fig:sft-loss-plot-12b}
    \end{subfigure}
    \hfill
    \begin{subfigure}[b]{0.48\textwidth}
        \centering
        \includegraphics[width=\textwidth]{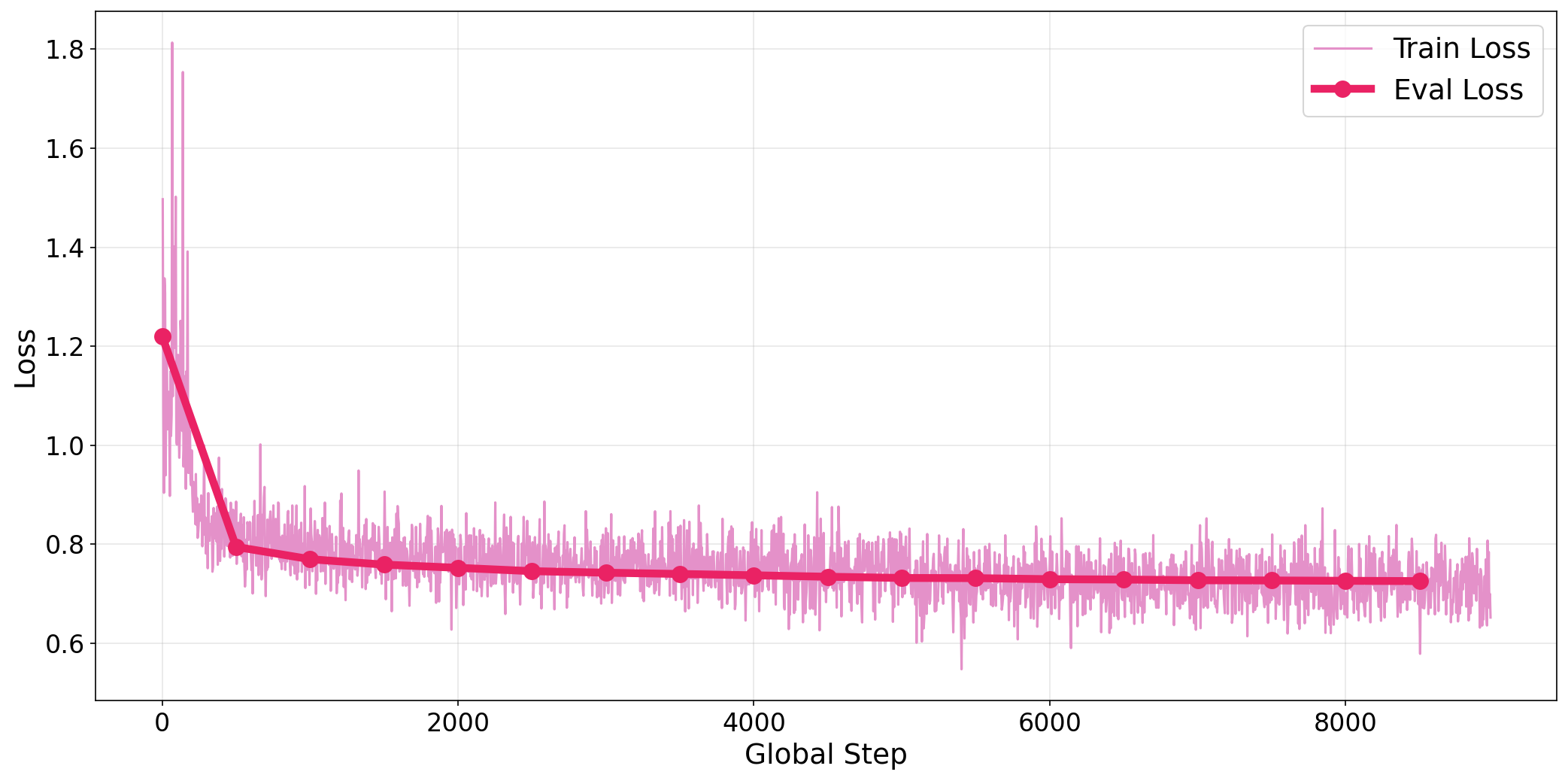}
        \caption{Soro pre-merge 27B}
        \label{fig:sft-loss-plot-27b}
    \end{subfigure}
    \caption{Training and evaluation loss during supervised instruction tuning of Soro Lite and Soro.}
    \label{fig:sft-combined-loss-plots}
\end{figure}

\paragraph{Convergence.} 
Figures~\ref{fig:sft-loss-plot-12b} and~\ref{fig:sft-loss-plot-27b} show smooth convergence of the training and validation loss during instruction tuning. Both models exhibit similar convergence behavior with no apparent overfitting.

\paragraph{Post-training Linear Merging.} 
\label{sec:merging}

\begin{figure}[hbpt!]
    \centering
    \includegraphics[width=0.6\textwidth]{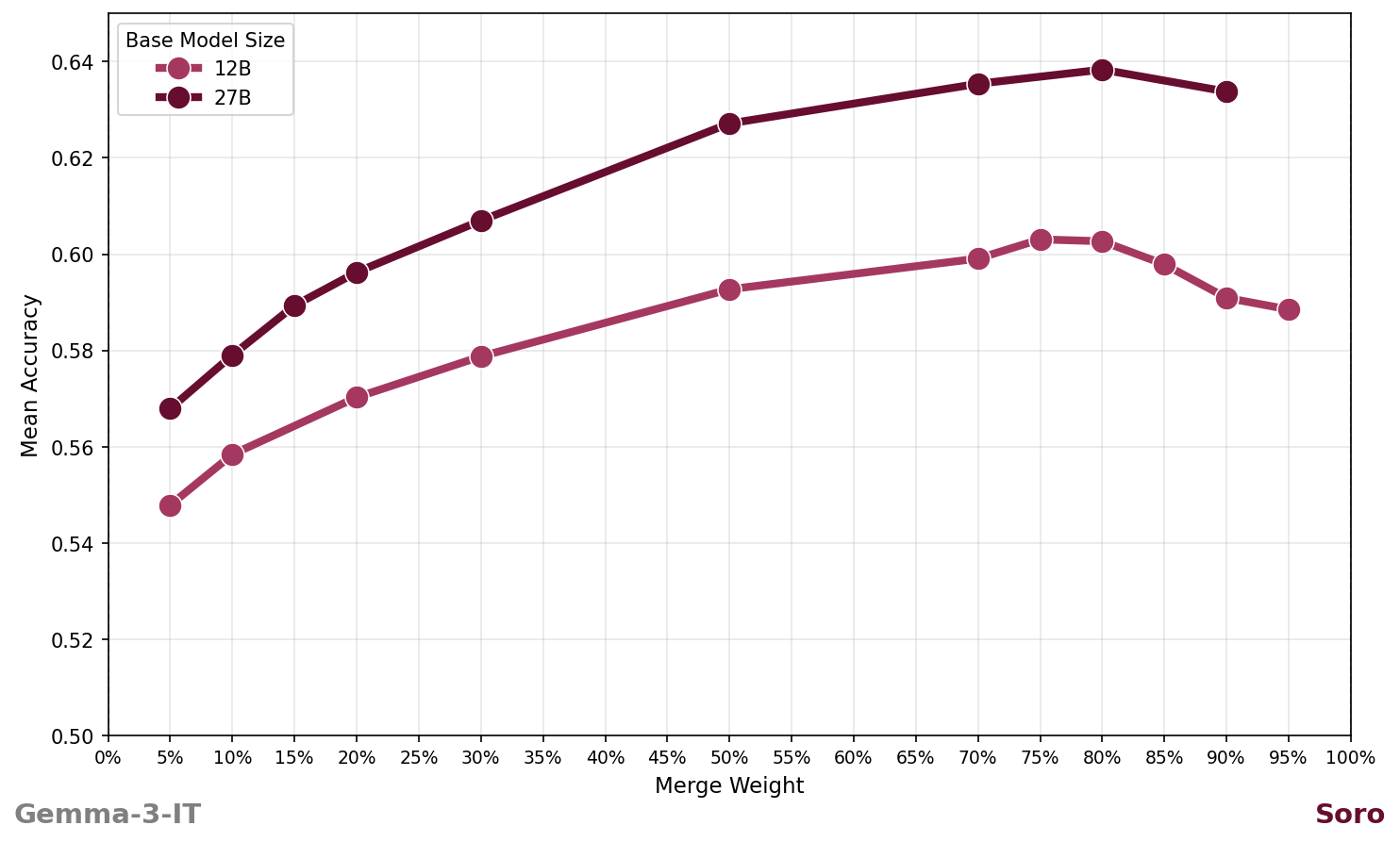}
    \caption{Impact of the linear merging weight on mean accuracy across Tajik benchmarks. The optimal ratio for both model sizes is approximately 80\% Soro / 20\% Gemma-3-IT.}
    \label{fig:merging}
\end{figure}

With instruction tuning complete, we experiment with linear merging of the open-source Gemma~3 instruction-tuned (IT) checkpoints and our Soro instruction-tuned checkpoints. The goal is to combine the cultural and linguistic specificity acquired during Tajik adaptation with the broader world knowledge retained in the Gemma~3-IT weights. We evaluate merged models across a range of interpolation ratios, varying the Gemma~3-IT contribution from 5\% to 90\%, and assess each variant on our Tajik benchmark suite. As shown in Figure~\ref{fig:merging}, the optimal mixture for both model sizes is approximately 20\% Gemma~3-IT and 80\% Soro: the Tajik-specialized weights should remain dominant, while a modest contribution from the original model helps recover general-purpose capabilities. We perform model merging using the \texttt{mergekit} library~\citep{goddard-etal-2024-arcees}. The final merged chatbots are referred to throughout as \textbf{Soro} and \textbf{Soro Lite}.

\section{Evaluation}
\label{sec:evaluation}

Having established the training and merging pipeline, we turn to evaluation. A central difficulty is the near-complete absence of established Tajik benchmarks; without them, it is impossible to measure whether adaptation is working or simply degrading general capabilities. We therefore developed a six-benchmark Tajik evaluation suite alongside the standard English MMLU benchmark, which serves as a capability retention check. Detailed benchmark descriptions are given in Appendix~\ref{sec:benchmarks}. Complete implementation details, including prompt templates in English and Tajik, are provided in Appendix~\ref{sec:eval_implementation}. 

\paragraph{General Benchmarks.}
The MMLU benchmark covers 57 subjects across STEM, humanities, and social sciences. We include it solely as a \emph{capability retention} benchmark: to verify that continual pretraining on Tajik data does not degrade the model's general English knowledge and reasoning.

\paragraph{Tajik-specific Benchmarks.}
We developed six Tajik-specific benchmarks covering linguistic competence, cultural knowledge, and academic subjects (Table~\ref{tab:tajik_benchmarks_overview}). These range from translations of established benchmarks (Tajik MMLU) to novel evaluations targeting Tajikistan-specific knowledge (Tajik-FactQA) and linguistic structure (TajLib).

\begin{table}[H]
\centering
\caption{Overview of Tajik-specific evaluation benchmarks. All benchmarks use a four-option multiple-choice format (A/B/C/D).}
\label{tab:tajik_benchmarks_overview}
\scalebox{0.82}{
\begin{tabularx}{1.21\textwidth}{lXXr}
\toprule
\textbf{Benchmark} & \textbf{Source} & \textbf{Description} & \textbf{Samples} \\
\midrule
Tajik MMLU (\S\ref{app:tajik-mmlu}) & Tajik translation of MMLU & General knowledge across 57 subjects (STEM, humanities, social sciences) & 1,525 \\
Tajik-FactQA (\S\ref{app:tajik-factual}) & Centralized University Entrance Exams & History, culture, geography, and national institutions of Tajikistan & 436 \\
Tajik History (\S\ref{app:tajik-history}) & School textbooks (grades 5--11) & National history curriculum from ancient civilizations to post-independence & 1,400 \\
Tajik Literature (\S\ref{app:tajik-literature}) & School textbooks (grades 5--11) & Classical and contemporary Tajik/Persian literature & 698 \\
TajLib (\S\ref{app:tajlib}) & National Testing Center Entrance Exams (2025) & Linguistic competence: phonetics, grammar, lexicology, orthography & 919 \\
Tajik Curated (\S\ref{app:tajik-curated}) & Diverse Tajik sources & Multi-domain: History, Biology, Literature, Law, Geography & 2,266 \\
\bottomrule
\end{tabularx}}
\end{table}

\subsection{Performance on Tajik Benchmarks}
\label{sec:tajik_results}

\begin{figure}[H]
    \centering
    \begin{tikzpicture}
    \begin{axis}[
        xbar,
        /pgf/bar shift=0pt,
        width=0.8\textwidth,
        height=6.5cm,
        xlabel={Accuracy (\%)},
        symbolic y coords={
            GPT-OSS-20B,
            Qwen2.5-7B-Instruct,
            Llama-3.1-8B-Instruct,
            Llama-3.2-11B-Vision-Instruct,
            Qwen3-8B,
            Qwen3-14B,
            Qwen3-32B,
            Gemma-3-12B-IT,
            Gemma-3-27B-IT,
            Soro Lite pre-merge,
            Soro Lite,
            Soro pre-merge,
            Soro
        },
        ytick=data,
        yticklabel style={font=\small},
        nodes near coords,
        nodes near coords align={horizontal},
        nodes near coords style={font=\footnotesize},
        every node near coord/.append style={
            /pgf/number format/.cd,
            fixed,
            precision=1,
            /tikz/.cd
        },
        xmin=25,
        xmax=70,
        bar width=0.3cm,
        enlarge y limits=0.055,
        axis line style={-},
        tick style={draw=none},
        ymajorgrids=false,
        xmajorgrids=true,
        grid style={dashed, gray!30},
    ]
    
    \addplot[draw=none, fill=none, forget plot] coordinates {
        (0,GPT-OSS-20B) (0,Qwen2.5-7B-Instruct) (0,Llama-3.1-8B-Instruct) (0,Llama-3.2-11B-Vision-Instruct) (0,Qwen3-8B) (0,Qwen3-14B) (0,Qwen3-32B) (0,Gemma-3-12B-IT) (0,Gemma-3-27B-IT) (0,Soro Lite pre-merge) (0,Soro Lite) (0,Soro pre-merge) (0,Soro)
    };
    
    \addplot[fill=othermodel, draw=othermodelborder] coordinates {
    (30.7,GPT-OSS-20B)
    };
    
    \addplot[fill=othermodel, draw=othermodelborder] coordinates {
    (34.4,Qwen2.5-7B-Instruct)
    (42.0,Qwen3-8B)
    (44.5,Qwen3-14B)
    (50.0,Qwen3-32B)
    };
    \addplot[fill=othermodel, draw=othermodelborder] coordinates {
    (39.0,Llama-3.1-8B-Instruct)
    (39.1,Llama-3.2-11B-Vision-Instruct)
    };
    \addplot[fill=othermodel, draw=othermodelborder] coordinates {
    (50.4,Gemma-3-12B-IT)
    (55.8,Gemma-3-27B-IT)
    };
    \addplot[fill=sorofamily, draw=sorofamilyborder] coordinates {
    (56.4,Soro Lite pre-merge)
    (62.3,Soro pre-merge)
    };
    \addplot[fill=soromain, draw=soromainborder] coordinates {
    (58.2,Soro Lite)
    (63.8,Soro)
    };
    
    \end{axis}
    \end{tikzpicture}
    \caption{Average accuracy on Tajik benchmarks across open-source models and full-precision Soro variants. Quantized results are in Section~\ref{sec:quantization}.}
    \label{fig:avg-tajik-performance}
\end{figure}
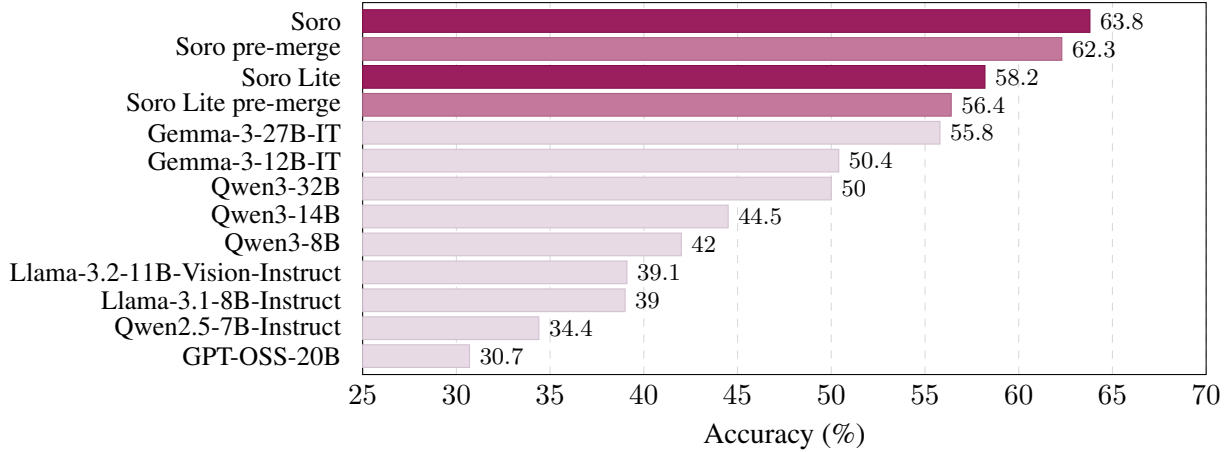

Our continual pretraining approach yields substantial improvements in Tajik language performance over baseline models of comparable size. Figure~\ref{fig:avg-tajik-performance} shows average accuracy across all Tajik-specific benchmarks (excluding English MMLU) for the evaluated open-source models and the final full-precision Soro models. Across both model sizes, Soro consistently outperforms its Gemma~3 base by roughly 6 percentage points on average, confirming that targeted continual pretraining improves both linguistic competence and culturally grounded knowledge even from a strong multilingual starting point. Soro models also achieve the best results among similarly sized open-source alternatives: Soro Lite (12B) surpasses Qwen3-14B, Qwen3-8B, and comparable Llama variants, while Soro (27B) outperforms Qwen3-32B despite having fewer parameters.

The gains are consistent across individual benchmarks and most pronounced on tasks requiring deeper linguistic and cultural understanding. On TajLib (Figure~\ref{fig:taj-lib-performance}), which evaluates Tajik linguistic competence, gains exceed 10 percentage points. Substantial improvements also appear on Tajik-FactQA (Figure~\ref{fig:taj-factual-performance}), Tajik History (Figure~\ref{fig:taj-history-grades-performance}), Tajik Literature (Figure~\ref{fig:taj-literature-performance}), and the general knowledge benchmarks, showing that adaptation improves both language understanding and culturally specific factual recall. English MMLU results (Figure~\ref{fig:mmlu-performance}) confirm that Tajik-focused adaptation does not substantially degrade broader reasoning and knowledge.

Merging further improves this trade-off. Linear interpolation between Soro-Instruct and the original Gemma~3-IT models at an 80\%/20\% ratio yields consistent gains of approximately 1.5--2 percentage points across both model sizes. The merged models preserve most of Soro's Tajik-specific improvements while closing most of the remaining gap in general English performance, suggesting that the knowledge acquired during Tajik adaptation and the knowledge retained in Gemma~3-IT are largely complementary. Per-benchmark results are provided in Appendix~\ref{sec:benchmarks}, and benchmark implementation details are in Appendix~\ref{sec:eval_implementation}.

\section{Inference}
\label{sec:inference}

Strong benchmark results are necessary but not sufficient: for Soro to be useful in Tajikistan's infrastructure, it must run efficiently on the limited hardware available at deployment sites. This section describes the quantization strategies applied to reduce model size and evaluates their effect on both Tajik performance and inference throughput. Throughout, ``full precision'' refers to the default BF16 (bfloat16) format used during training.

\subsection{Quantization}
\label{sec:quantization} 

We quantize the final merged models, Soro Lite and Soro, using two complementary approaches: dynamic FP8 quantization~\citep{kuzmin2024fp8quantizationpowerexponent} and GPTQ INT4 quantization~\citep{frantar2023gptqaccurateposttrainingquantization}. Both are applied using the \texttt{llmcompressor} library~\citep{llmcompressor2024}. Detailed quantization configurations are provided in Appendix~\ref{sec:quantization_details}.

Table~\ref{tab:quant_model_comparison} presents the performance of quantized models across our Tajik evaluation suite alongside their full-precision counterparts. Both methods preserve strong Tajik-language capabilities with minimal accuracy degradation. FP8 quantization reduces memory footprint by approximately 50\% with negligible accuracy loss; for the 27B model, FP8 performance remains effectively identical to the full-precision result. GPTQ INT4 achieves $4\times$ memory compression while retaining the majority of the merged models' performance gains. Both quantized variants still substantially outperform their respective Gemma~3 baselines. We attribute this robustness in part to the Gemma~3 base weights, which were trained with quantization-aware training (QAT)~\citep{gemmateam2025gemma3technicalreport}, a property that appears to be preserved through our LoRA-based adaptation and linear merge with Gemma~3-IT.

\begin{table}[H]
\centering
\caption{Average accuracy on Tajik benchmarks (\%). Default precision is BF16. Gemma-3-IT and full-precision Soro columns are included as reference baselines.}
\label{tab:quant_model_comparison}
\scalebox{0.82}{
\begin{tabular}{l|ccccc}
\toprule
\textbf{Size} & \textbf{Gemma-3-IT} & \textbf{Soro pre-merge} & \textbf{Soro} & \textbf{Soro FP8} & \textbf{Soro GPTQ INT4} \\
\midrule
\textbf{12B} & 50.4 & 56.4 & 58.2 & 57.7 & 56.5 \\
\textbf{27B} & 55.8 & 62.3 & 63.8 & 63.9 & 62.3 \\
\midrule
\end{tabular}}
\end{table}

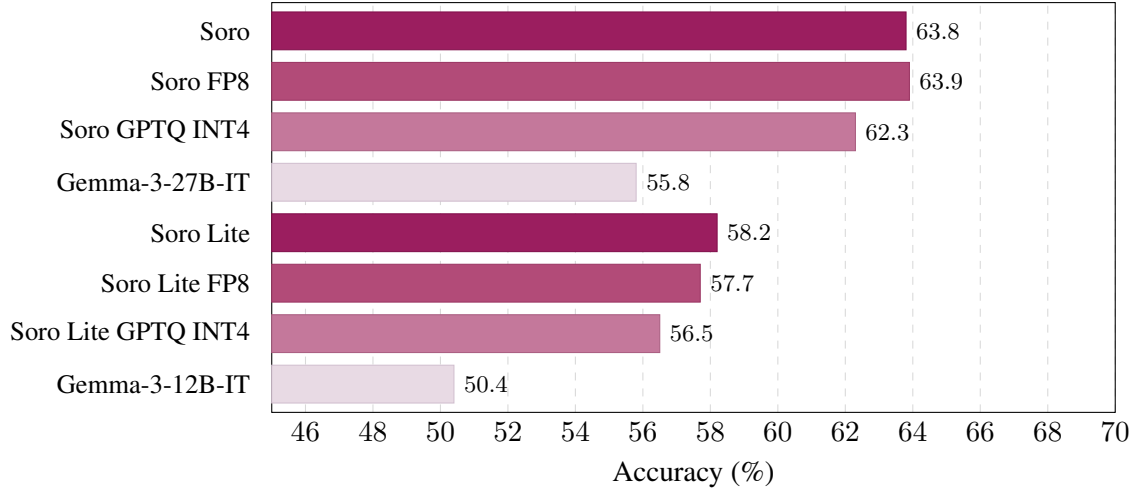
\begin{figure}[H]
    \centering
    \begin{tikzpicture}
    \begin{axis}[
        xbar,
        /pgf/bar shift=0pt,
        width=0.8\textwidth,
        height=7cm,
        xlabel={Accuracy (\%)},
        symbolic y coords={
            Gemma-3-12B-IT,
            Soro Lite GPTQ INT4,
            Soro Lite FP8,
            Soro Lite,
            Gemma-3-27B-IT,
            Soro GPTQ INT4,
            Soro FP8,
            Soro
        },
        ytick=data,
        yticklabel style={font=\small},
        nodes near coords,
        nodes near coords align={horizontal},
        nodes near coords style={font=\footnotesize},
        every node near coord/.append style={
            /pgf/number format/.cd,
            fixed,
            precision=1,
            /tikz/.cd
        },
        xmin=45,
        xmax=70,
        bar width=0.5cm,
        enlarge y limits=0.08,
        axis line style={-},
        tick style={draw=none},
        ymajorgrids=false,
        xmajorgrids=true,
        grid style={dashed, gray!30},
    ]
    
    \addplot[draw=none, fill=none, forget plot] coordinates {
        (0,Gemma-3-12B-IT) (0,Soro Lite GPTQ INT4) (0,Soro Lite FP8) (0,Soro Lite)
        (0,Gemma-3-27B-IT) (0,Soro GPTQ INT4) (0,Soro FP8) (0,Soro)
    };

    \addplot[fill=othermodel, draw=othermodelborder] coordinates {
        (50.4,Gemma-3-12B-IT)
        (55.8,Gemma-3-27B-IT)
    };

    \addplot[fill=sorofamily, draw=sorofamilyborder] coordinates {
        (56.5,Soro Lite GPTQ INT4)
        (62.3,Soro GPTQ INT4)
    };

    \addplot[fill=sorofp8, draw=sorofp8border] coordinates {
        (57.7,Soro Lite FP8)
        (63.9,Soro FP8)
    };

    \addplot[fill=soromain, draw=soromainborder] coordinates {
        (58.2,Soro Lite)
        (63.8,Soro)
    };
    
    \end{axis}
    \end{tikzpicture}
    \caption{Average accuracy on Tajik benchmarks for quantized Soro variants compared to full-precision Soro and Gemma~3 baselines. FP8 quantization preserves near-identical performance while halving memory; GPTQ INT4 achieves 4$\times$ compression with modest accuracy trade-offs. Full results across all models are reported in Figure~\ref{fig:avg-tajik-performance}.}
    \label{fig:avg-tajik-performance-quant}
\end{figure}

Figure~\ref{fig:avg-tajik-performance-quant} shows the average accuracy of the quantized Soro variants on Tajik benchmarks. Both FP8 and GPTQ INT4 preserve the large majority of the Tajik-specific gains reported in Section~\ref{sec:tajik_results}, with FP8 showing negligible degradation and INT4 retaining most improvements despite $4\times$ memory reduction.

\begin{table}[h]
\centering
\caption{Model memory footprint for quantized variants.}
\label{tab:memory-footprint}
\scalebox{0.9}{
\begin{tabular}{lccc}
\toprule
\textbf{Model Variant} & \textbf{Precision} & \textbf{Memory (GB)} & \textbf{BF16 Baseline (GB)} \\
\midrule
Soro Lite FP8 & FP8 & 13.0 & 24.0 \\
Soro Lite GPTQ INT4 & INT4 & 6.6 & 24.0 \\
Soro FP8 & FP8 & 29.0 & 54.0 \\
Soro GPTQ INT4 & INT4 & 13.5 & 54.0 \\
\bottomrule
\end{tabular}}
\end{table}

\subsection{Hardware Deployment}
 \label{sec:hardware-deployment}

The quantized model sizes enable deployment across a range of consumer and enterprise GPUs. Soro Lite GPTQ INT4 requires only 6.6~GB and can run on widely available consumer GPUs such as the NVIDIA RTX 3060 (12~GB) or RTX 4060 (8~GB). The FP8 variant at 13~GB fits on an RTX 4090 (24~GB). For the full 27B model, the GPTQ INT4 variant at 13.5~GB enables single-GPU deployment on similar hardware, while the FP8 variant at 29~GB requires a higher-end card such as the A100 (40~GB or 80~GB). These memory reductions are critical for deployment in Tajikistan's educational infrastructure, where data-center-class hardware is scarce.

\section{Real-World Deployment Case Study}
\label{sec:deployment}

The quantization strategies described above are not merely an engineering convenience: they directly determine whether Soro can reach the schools and government offices it was designed for. Soro is currently undergoing a structured pilot deployment in Tajikistan's education sector, coordinated by the Artificial Intelligence Council (AIC) under the Ministry of Industry and New Technologies, in collaboration with UNICEF Tajikistan and with support from the Ministry of Education and Science. This section describes the deployment setting, the use cases under evaluation, and preliminary observations on infrastructure and user experience.

\subsection{Pilot Overview}
The pilot initially encompassed 30 schools in Dushanbe, reaching 770 teachers and over 2,000 students through two complementary channels. Teachers access Soro via integration with the existing national educational platform eDonish, through a partnership with Smart City Dushanbe. Students interact with the model during dedicated ``Introduction to AI'' lessons, where Soro serves as a hands-on learning tool alongside the regular curriculum. The pilot is part of the broader \textit{Project Soro}\footnote{See the official UNICEF Tajikistan press release: \href{https://www.unicef.org/tajikistan/press-releases/tajikistan-launches-project-soro}{unicef.org/tajikistan}.} initiative, which aims to deploy AI-powered tools across the national education system within the framework of Tajikistan's National AI Strategy (NAIS-2040).

\begin{figure}[H]
     \centering
     \begin{subfigure}[b]{0.4\textwidth}
         \centering
         \includegraphics[width=\linewidth]{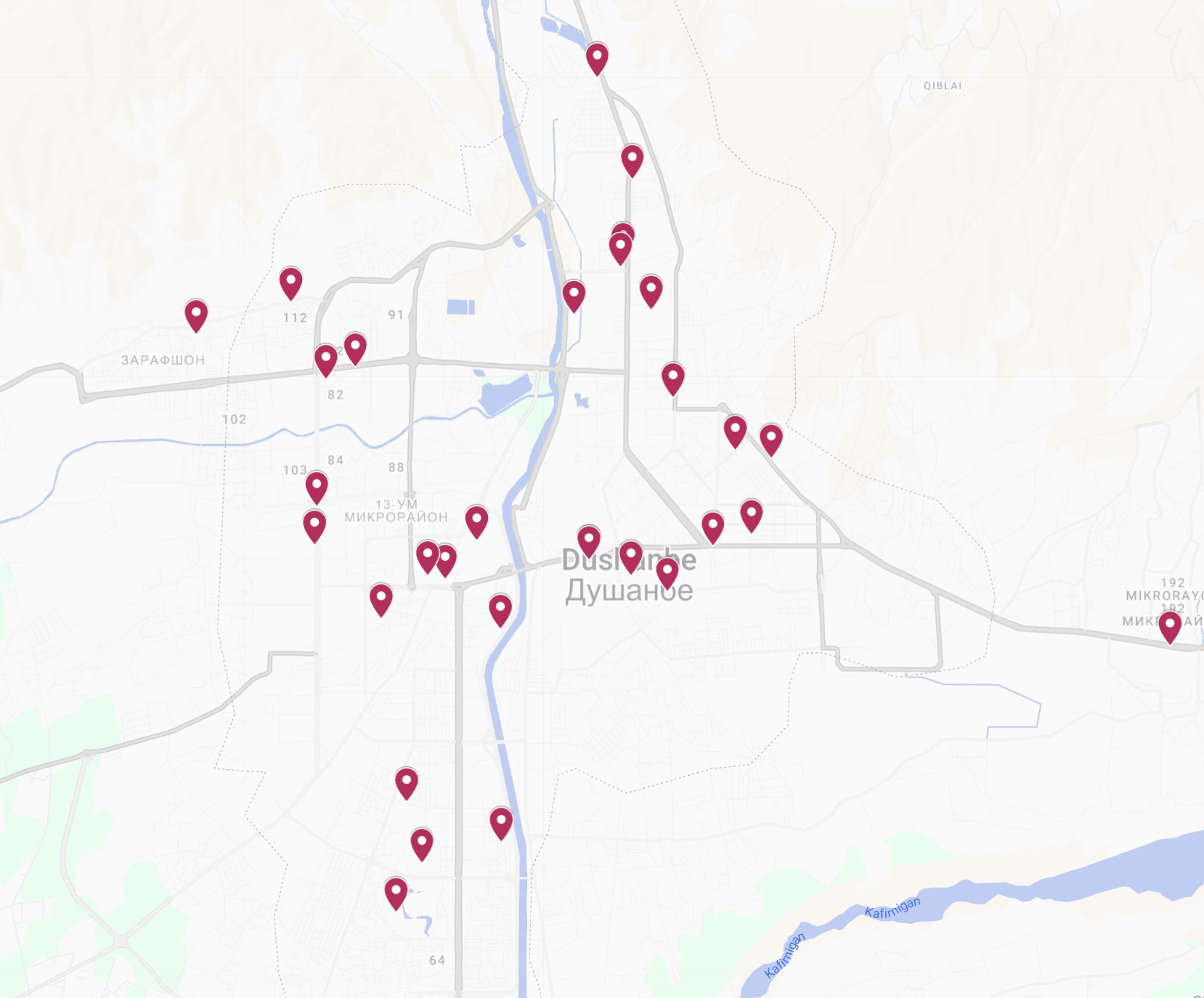}
         \caption{\raggedright Initial 30-school pilot deployment in Dushanbe. Blue markers indicate the schools.}
         \label{fig:map_dushanbe}
     \end{subfigure}
     \hfill
     \begin{subfigure}[b]{0.52\textwidth}
         \centering
         \includegraphics[width=\linewidth]{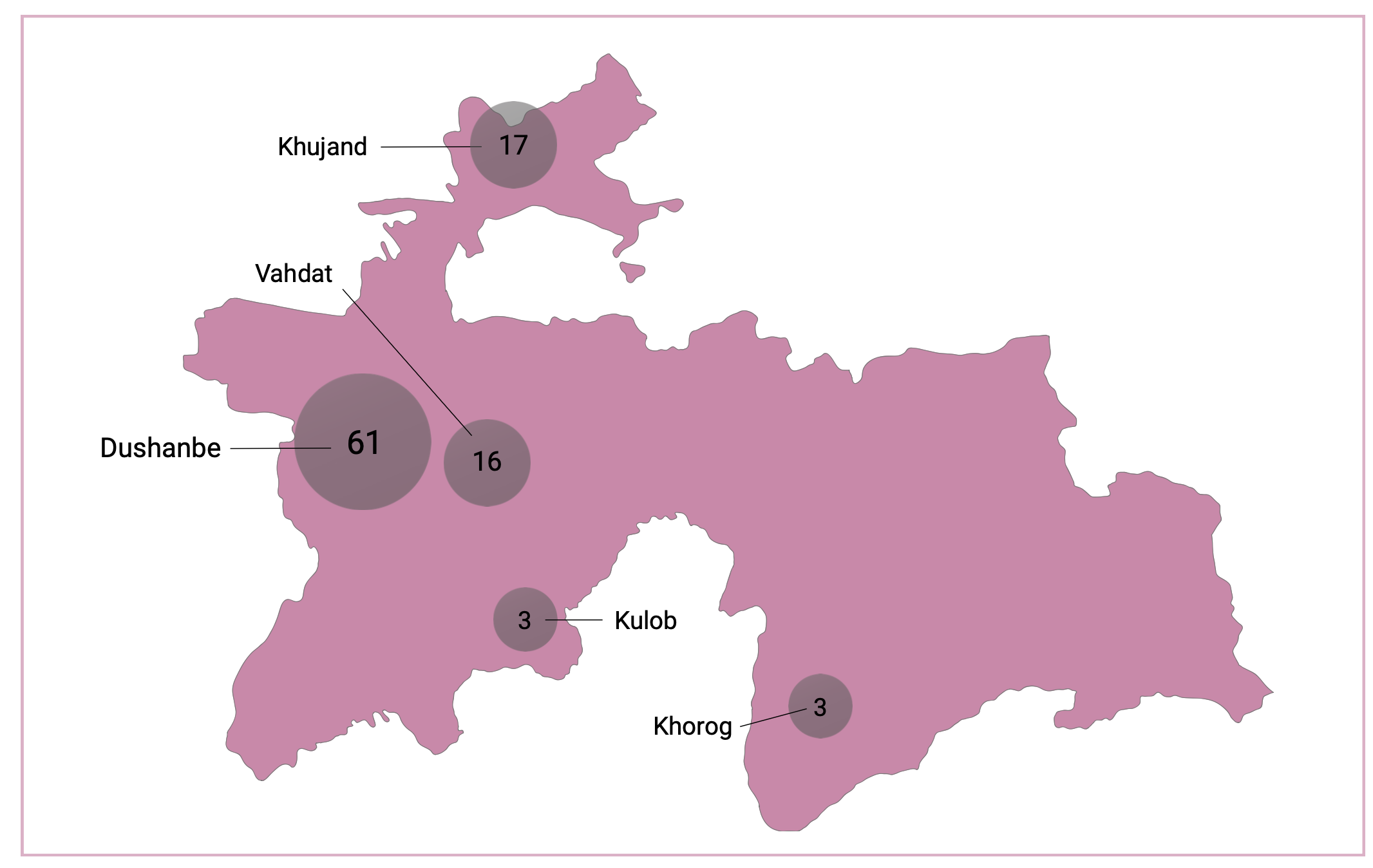}
         \caption{\raggedright Current national expansion across Tajikistan (100 schools).}
         \label{fig:map_national}
     \end{subfigure}
     
     \caption{Geographic distribution of Soro deployments. (a) Initial 30-school pilot in Dushanbe; (b) Current national expansion to 100 schools across Dushanbe (61), Khujand (17), Vahdat (16), Kulob (3), and Khorog (3). The long-term NAIS-2040 target is 4,000+ schools by the 2027--28 academic year.}
     \label{fig:soro-deployment-maps}
\end{figure}

The deployment has since expanded to 100 schools across five cities: Dushanbe (61 schools), Khujand (17), Vahdat (16), Kulob (3), and Khorog (3). This expansion extends coverage beyond the capital to include Sughd province (via Khujand, the country's second-largest city), the Districts of Republican Subordination (via Vahdat), Khatlon province (via Kulob), and the Gorno-Badakhshan Autonomous Region (via Khorog), ensuring representation across all major administrative regions of Tajikistan. Figure~\ref{fig:map_dushanbe} shows the initial 30-school deployment in Dushanbe; Figure~\ref{fig:map_national} presents the current national distribution across 100 schools.

The roadmap follows a phased scaling trajectory aligned with NAIS-2040: 500 schools by the end of 2026, with an ultimate target of integrating Soro into all 4,000+ schools nationwide by the 2027--28 academic year. This progression is designed to proceed in parallel with the national rollout of this AI curriculum, so that infrastructure deployment and pedagogical adoption advance together. The edge deployment capabilities enabled by our quantized model variants (Section~\ref{sec:quantization}) are essential to this trajectory: schools in Khorog, Kulob, and other regional centers face intermittent connectivity that makes reliance on centralized API infrastructure alone infeasible.

\subsection{Use Cases}

Three primary use cases are being evaluated during the pilot.

\textbf{Personalized learning assistance.} Students interact with Soro in Tajik to ask follow-up questions on lesson material, request simplified explanations, and study at their own pace. This use case is especially relevant in Tajikistan, where traditional classroom dynamics can discourage students, particularly in rural and underserved communities, from asking questions or admitting confusion. A private, native-language interface removes some of that friction and can serve as a complementary resource both in class and at home.

\textbf{Instructor support and learning diagnostics.} Teachers use Soro to generate lesson summaries, draft instructional guidance, and identify patterns in student comprehension difficulties. In settings where class sizes are large and per-student attention is limited, this functionality is designed to help educators identify where students are falling behind and adjust lesson plans accordingly. Early feedback indicates measurable time savings in lesson preparation.

\textbf{AI literacy curriculum support.} The pilot operates alongside Tajikistan's national rollout of ``Introduction to AI'' as a mandatory school subject, a key pillar of NAIS-2040 targeting 4,000 schools and 2 million students by the 2027--28 academic year. A recently completed pilot of this curriculum across 25 schools and 275 students yielded an 80\% post-assessment pass rate, with an average score of 18 out of 20 points. Soro's presence in classrooms reinforces this curriculum by giving students direct, hands-on exposure to a locally developed AI system and practical familiarity with language model capabilities and limitations.

Representative sample interactions illustrating the model's behavior on curriculum-relevant educational queries are provided in Appendix~\ref{app:sample-interactions}.

\subsection{Infrastructure and Connectivity}

Infrastructure conditions vary considerably across deployment sites, shaping our hosting strategy.

\textbf{Urban deployment.} The majority of pilot schools are located in Dushanbe and other major cities, where internet connectivity is stable and sufficient for centralized model hosting. Soro (27B) is served from on-premise infrastructure within Tajikistan, with no major outages or service interruptions reported during the pilot period. However, teacher feedback indicates that response speed can still be too slow for some live classroom interactions, even when the system is available and functioning reliably.

\textbf{Rural scalability challenges.} The upcoming expansion to 500 schools will include institutions in remote and rural regions of Tajikistan, where connectivity is intermittent or unreliable. Preliminary site assessments indicate that centralized hosting alone will not be sufficient for nationwide coverage. We are therefore evaluating edge deployment strategies in which quantized versions of Soro (see Section~\ref{sec:quantization}) would run locally on modest school hardware, reducing dependence on continuous network connectivity.

\subsection{Preliminary Observations}

A comprehensive user study remains planned for future work. To complement the early qualitative observations reported here, Appendix~\ref{app:teacher-survey} presents results from a structured teacher feedback survey conducted during the pilot deployment, including quantitative ratings and a thematic analysis of open-ended responses. Several qualitative observations have emerged from the initial deployment:

\begin{enumerate}[label=(\roman*), leftmargin=2em]
    \item Students report increased willingness to engage with lesson material when interacting with Soro compared to traditional classroom settings, consistent with the hypothesis that a private, native-language interface reduces social barriers to learning.
    \item Teachers note time savings in lesson preparation and value the availability of Tajik-language AI assistance, which was previously unavailable from any commercial or open-source system.
    \item The model occasionally produces factually incorrect responses on highly specific local topics (e.g., regional geography, recent institutional changes), pointing to the need for continued expansion of the instruction tuning dataset with verified local knowledge.
    \item Some teachers have requested finer control over response complexity to match different grade levels, motivating the development of grade-adapted system prompts for future iterations.
\end{enumerate}

\noindent These observations are actively informing ongoing improvements to dataset composition and deployment-specific inference configurations.

\section{Ethical Considerations}
\label{sec:ethical-considerations}

Deploying a Tajik-specialized LLM in schools and public institutions raises a set of practical ethical concerns that we take seriously throughout the development and deployment process.

Our training corpus aggregates publicly available web and document sources, which may contain personal information, copyrighted material, or other sensitive content. Filtering, deduplication, and quality-control procedures reduce but do not eliminate these risks; data provenance and privacy require continuous auditing.

The model may also reproduce social, cultural, or political biases present in its training data. Because we aggregate open-web resources, textbooks, and historical books, and additionally merge the adapted model weights with those of Gemma~3, maintaining full control over the worldview reflected in the final model is not feasible. The resulting system may combine perspectives from older generations in Tajikistan, the contemporary educational system, online discourse within and beyond the country, and the prior training distribution of Gemma~3. Managing these layered influences remains an ongoing challenge.

As with all language models, Soro may generate inaccurate or hallucinated responses. This risk is especially significant in educational settings involving underage students who may treat the system as authoritative. To mitigate this, we implement inference-time output filtering to reduce inappropriate content (sexual content, self-harm, and hate or harassment) and limit misleading generations. Even so, our Tajik-specific benchmarks measure performance on structured tasks and do not guarantee reliability in open-ended settings; deployment interfaces should encourage verification and human oversight. Generative models are inherently dual-use and could be misused for misinformation or social engineering in Tajik.

Our underlying goal is to expand access to information for populations in remote areas who may not be comfortable using English, and to give students in underserved regions access to modern AI tools for learning. This work is a step in that direction, and we remain committed to improving the model based on real-world deployment findings while remaining mindful of the responsibilities that come with deployment at scale.


\section{Conclusion}

We have presented Soro, a family of lightweight, Tajik-specialized conversational language models built for deployment under the compute, memory, and connectivity constraints typical of educational and public-sector settings in Tajikistan. Starting from open-weight Gemma~3 checkpoints, we performed targeted continual pretraining on a curated 1.9-billion-token Tajik-only corpus, followed by supervised instruction tuning on 40K teacher-style examples. The resulting models, Soro (27B) and Soro Lite (12B), deliver substantial gains in Tajik linguistic competence, cultural knowledge, and curriculum-aligned factual recall while largely preserving English and general multilingual capabilities.

Because Tajik has almost no existing evaluation benchmarks, we constructed a suite of Tajik-specific multiple-choice datasets and we make them publicly available on Hugging Face. Across this suite, the final Soro variants achieve average accuracy improvements of 8--13\% over same-size Gemma~3 instruction-tuned baselines and outperform all other openly available models of comparable size, frequently by large margins on culturally and linguistically specific tasks.

The ongoing education-sector pilot, currently active across 100 schools in five regions of Tajikistan, represents, to the best of our knowledge, one of the first structured, government-backed deployments of a locally adapted LLM in a low-resource educational system. Early qualitative feedback highlights increased student engagement, time savings for teachers, and the value of a system that communicates fluently in curriculum-aligned Tajik.

Several directions offer the most traction for future improvement. Expanding high-quality, community-verified Tajik instruction and preference data, ideally through partnerships with educators, would help reduce the factual errors identified during the pilot. Adding native multimodal capabilities would allow the model to engage with Tajik educational materials such as diagrams, maps, and textbook illustrations. Further refining Tajik-specific tokenization would likely improve efficiency, and retrieval-augmented generation tied to a local knowledge base could substantially reduce errors on recent or hyper-local topics.

Soro demonstrates that focused, resource-conscious adaptation of a strong open foundation model can produce high-quality, deployable language technology for a severely under-resourced language. Beyond Tajik, the methodology and deployment lessons described here are directly transferable to other low-resource languages facing similar infrastructure constraints. We intend this work as both a practical blueprint for such efforts and a concrete contribution to the educational opportunities available to students and teachers across Tajikistan.

\section*{Acknowledgements}

We extend our sincere gratitude to the Artificial Intelligence Council (AIC) under the Ministry of Industry and New Technologies of the Republic of Tajikistan for their overall coordination of the education pilot and for supporting data collection efforts. We are particularly grateful to Suhrob Rajabov for his strategic guidance throughout the project. Special thanks go to Temur Nazarov and  Gulsum Kosumbekova for their extensive work in organizing the data collection process, and to Nozanin Kasymova at the AIC for her valuable support. We also thank the Ministry of Industry and New Technologies for their institutional backing of the project within the framework of the National AI Strategy (NAIS-2040).

We acknowledge the Smart City State Unitary Enterprise (Smart City Dushanbe) for providing access to the eDonish platform, their assistance with data sourcing, and their role in organizing teacher training sessions and fostering model adoption across the participating schools.

We thank the Ministry of Education and Science of the Republic of Tajikistan for their institutional support, facilitation of access to national educational materials, and assistance with curriculum-aligned data sourcing. We also thank the Main Department of Education of the City of Dushanbe for their operational support throughout the pilot, including providing venues for training sessions and feedback collection events and enabling teachers to test and evaluate Soro in practice.

We acknowledge the United Nations Children's Fund Tajikistan (UNICEF Tajikistan) for their mentorship and expert guidance on child safety, age-appropriate content, and responsible deployment considerations relevant to educational settings involving minors.

We thank zypl.ai, our technical and strategic partner on this initiative, for their contributions across data collection and formatting, the promotion and deployment of Soro, and the coordination of partnerships and integrations with collaborating organizations.

Furthermore, we are grateful to the National Library of Tajikistan and the Academy of Sciences of Tajikistan for providing access to archival and reference materials.

Finally, we thank the 100 pilot schools across Dushanbe, Khujand, Vahdat, Kulob, and Khorog, and the teachers and students who participated in the deployment and provided the feedback that informs this research.

\clearpage
\bibliographystyle{plainnat}
\bibliography{references}

\appendix
\clearpage

\section{Benchmarks}
\label{sec:benchmarks}

This section provides detailed descriptions of the Tajik-specific evaluation benchmarks introduced in this work. All benchmarks use a four-option multiple-choice format (A/B/C/D) and are evaluated using the logit-based probability extraction procedure described in Appendix~\ref{sec:eval_implementation}. Unless otherwise noted, all questions and answer options are written entirely in Tajik.

\subsection{MMLU (English)}
\label{app:mmlu-en}

The Massive Multitask Language Understanding (MMLU) benchmark~\citep{hendryckstest2021} covers 57 subjects across STEM, humanities, social sciences, and professional domains. We use the standard HuggingFace dataset\footnote{\url{https://huggingface.co/datasets/cais/mmlu}} and include it as a \textit{capability retention} benchmark: its sole purpose is to verify that continual pretraining on Tajik data does not degrade general-purpose English knowledge and reasoning. MMLU is the only English-language benchmark in our evaluation suite; all remaining benchmarks are in Tajik.

\begin{figure}[H]
    \centering
    \begin{tikzpicture}
    \begin{axis}[
        xbar,
        /pgf/bar shift=0pt,
        width=0.8\textwidth,
        height=8cm,
        xlabel={Accuracy (\%)},
        symbolic y coords={
            Llama-3.2-11B-Vision-Instruct,
            GPT-OSS-20B,
            Llama-3.1-8B-Instruct,
            Soro Lite pre-merge,
            Qwen2.5-7B-Instruct,
            Soro Lite GPTQ INT4,
            Soro Lite FP8,
            Soro Lite,
            Qwen3-8B,
            Gemma-3-12B-IT,
            Soro pre-merge,
            Qwen3-14B,
            Soro GPTQ INT4,
            Soro,
            Soro FP8,
            Gemma-3-27B-IT,
            Qwen3-32B
        },
        ytick=data,
        yticklabel style={font=\small},
        nodes near coords,
        nodes near coords align={horizontal},
        nodes near coords style={font=\footnotesize},
        every node near coord/.append style={
            /pgf/number format/.cd,
            fixed,
            precision=1,
            /tikz/.cd
        },
        xmin=25,
        xmax=100,
        bar width=0.3cm,
        enlarge y limits=0.04,
        axis line style={-},
        tick style={draw=none},
        ymajorgrids=false,
        xmajorgrids=true,
        grid style={dashed, gray!30},
    ]
    
    \addplot[draw=none, fill=none, forget plot] coordinates {
        (0,Llama-3.2-11B-Vision-Instruct) (0,GPT-OSS-20B) (0,Llama-3.1-8B-Instruct) (0,Soro Lite pre-merge) (0,Qwen2.5-7B-Instruct) (0,Soro Lite GPTQ INT4) (0,Soro Lite FP8) (0,Soro Lite) (0,Qwen3-8B) (0,Gemma-3-12B-IT) (0,Soro pre-merge) (0,Qwen3-14B) (0,Soro GPTQ INT4) (0,Soro) (0,Soro FP8) (0,Gemma-3-27B-IT) (0,Qwen3-32B)
    };
    
    \addplot[fill=othermodel, draw=othermodelborder] coordinates {
        (64.3,Llama-3.2-11B-Vision-Instruct)
        (65.8,Llama-3.1-8B-Instruct)
    };
    
    \addplot[fill=othermodel, draw=othermodelborder] coordinates {
        (65.0,GPT-OSS-20B)
    };
    
    \addplot[fill=sorofamily, draw=sorofamilyborder] coordinates {
        (68.5,Soro Lite pre-merge)
        (75.7,Soro pre-merge)
    };
    
    \addplot[fill=othermodel, draw=othermodelborder] coordinates {
        (70.1,Qwen2.5-7B-Instruct)
        (72.8,Qwen3-8B)
        (76.6,Qwen3-14B)
        (80.6,Qwen3-32B)
    };
    
    \addplot[fill=sorofamily, draw=sorofamilyborder] coordinates {
        (70.7,Soro Lite GPTQ INT4)
        (76.7,Soro GPTQ INT4)
    };
    
    \addplot[fill=sorofp8, draw=sorofp8border] coordinates {
        (72.3,Soro Lite FP8)
        (78.2,Soro FP8)
    };
    
    \addplot[fill=soromain, draw=soromainborder] coordinates {
        (72.6,Soro Lite)
        (78.2,Soro)
    };
    
    \addplot[fill=othermodel, draw=othermodelborder] coordinates {
        (74.1,Gemma-3-12B-IT)
        (78.6,Gemma-3-27B-IT)
    };
    
    \end{axis}
    \end{tikzpicture}
    \caption{MMLU Benchmark}
    \label{fig:mmlu-performance}
\end{figure}
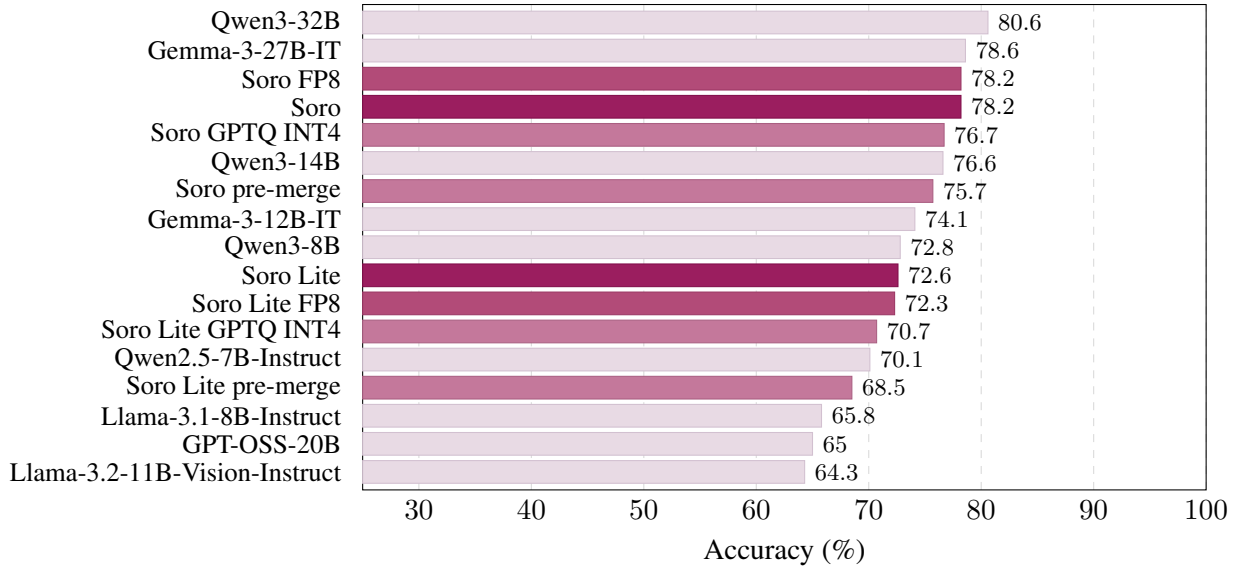

\subsection{Tajik Curated General Benchmark}
\label{app:tajik-curated}

The Tajik Curated General Benchmark comprises 2{,}266 questions distributed across six domains: History, Biology, Literature, Tajik Language (grammar, morphology, and syntax), Law, and Geography. The questions were manually curated from diverse Tajik-language sources to assess cross-domain breadth rather than depth in any single area. Its defining feature is thematic diversity: while the dedicated benchmarks (Appendices~\ref{app:tajik-factual}--\ref{app:tajlib}) probe individual domains in depth, the Curated General Benchmark tests whether a model maintains balanced competence across multiple fields simultaneously. The Tajik Language questions target grammatical and morphological competence; the Law subset covers constitutional and civil law topics relevant to Tajikistan's legal framework. This dataset is available on Hugging Face.\footnote{\url{https://huggingface.co/datasets/zehnlab/tajik-curated-benchmark}}

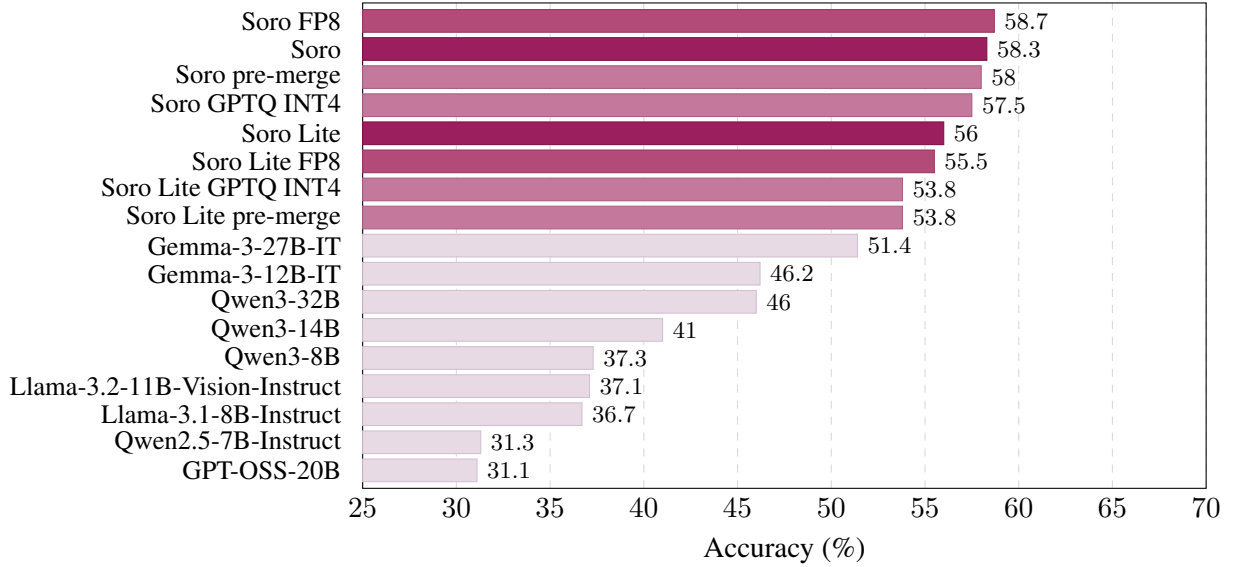
\begin{figure}[H]
    \centering
    \begin{tikzpicture}
    \begin{axis}[
        xbar,
        /pgf/bar shift=0pt,
        width=0.8\textwidth,
        height=8cm,
        xlabel={Accuracy (\%)},
        symbolic y coords={
            GPT-OSS-20B,
            Qwen2.5-7B-Instruct,
            Llama-3.1-8B-Instruct,
            Llama-3.2-11B-Vision-Instruct,
            Qwen3-8B,
            Qwen3-14B,
            Qwen3-32B,
            Gemma-3-12B-IT,
            Gemma-3-27B-IT,
            Soro Lite pre-merge,
            Soro Lite GPTQ INT4,
            Soro Lite FP8,
            Soro Lite,
            Soro GPTQ INT4,
            Soro pre-merge,
            Soro,
            Soro FP8
        },
        ytick=data,
        yticklabel style={font=\small},
        nodes near coords,
        nodes near coords align={horizontal},
        nodes near coords style={font=\footnotesize},
        every node near coord/.append style={
            /pgf/number format/.cd,
            fixed,
            precision=1,
            /tikz/.cd
        },
        xmin=25,
        xmax=70,
        bar width=0.3cm,
        enlarge y limits=0.04,
        axis line style={-},
        tick style={draw=none},
        ymajorgrids=false,
        xmajorgrids=true,
        grid style={dashed, gray!30},
    ]
    
    \addplot[draw=none, fill=none, forget plot] coordinates {
        (0,GPT-OSS-20B) (0,Qwen2.5-7B-Instruct) (0,Llama-3.1-8B-Instruct) (0,Llama-3.2-11B-Vision-Instruct) (0,Qwen3-8B) (0,Qwen3-14B) (0,Qwen3-32B) (0,Gemma-3-12B-IT) (0,Gemma-3-27B-IT) (0,Soro Lite pre-merge) (0,Soro Lite GPTQ INT4) (0,Soro Lite FP8) (0,Soro Lite) (0,Soro GPTQ INT4) (0,Soro pre-merge) (0,Soro) (0,Soro FP8)
    };
    
    \addplot[fill=othermodel, draw=othermodelborder] coordinates {
        (31.1,GPT-OSS-20B)
    };
    
    \addplot[fill=othermodel, draw=othermodelborder] coordinates {
        (31.3,Qwen2.5-7B-Instruct)
        (37.3,Qwen3-8B)
        (41.0,Qwen3-14B)
        (46.0,Qwen3-32B)
    };
    
    \addplot[fill=othermodel, draw=othermodelborder] coordinates {
        (36.7,Llama-3.1-8B-Instruct)
        (37.1,Llama-3.2-11B-Vision-Instruct)
    };
    
    \addplot[fill=othermodel, draw=othermodelborder] coordinates {
        (46.2,Gemma-3-12B-IT)
        (51.4,Gemma-3-27B-IT)
    };
    
    \addplot[fill=sorofamily, draw=sorofamilyborder] coordinates {
        (53.8,Soro Lite pre-merge)
        (58.0,Soro pre-merge)
    };
    
    \addplot[fill=sorofamily, draw=sorofamilyborder] coordinates {
        (53.8,Soro Lite GPTQ INT4)
        (57.5,Soro GPTQ INT4)
    };
    
    \addplot[fill=sorofp8, draw=sorofp8border] coordinates {
        (55.5,Soro Lite FP8)
        (58.7,Soro FP8)
    };
    
    \addplot[fill=soromain, draw=soromainborder] coordinates {
        (56.0,Soro Lite)
        (58.3,Soro)
    };
    
    \end{axis}
    \end{tikzpicture}
    \caption{Tajik Curated Benchmark}
    \label{fig:taj-curated-performance}
\end{figure}

\subsection{Tajik Factual Benchmark (Tajik-FactQA)}
\label{app:tajik-factual}

Tajik-FactQA consists of 436 multiple-choice questions drawn from the Centralized University Entrance Examinations (\textit{Markazi millii testī}) administered to high-school graduates seeking university admission in Tajikistan. The questions assess knowledge of Tajikistan's history, culture, geography, and national institutions. Because the source material reflects officially standardized examination content, this benchmark provides a controlled measure of culturally grounded factual recall at a post-secondary difficulty level. This dataset is available on
Hugging Face.\footnote{\url{https://huggingface.co/datasets/zehnlab/Tajik-FactQA-benchmark}}

\paragraph{Distinction from Tajik History Benchmark.} 
Although both this benchmark and the Tajik History Benchmark (Appendix~\ref{app:tajik-history}) involve historical knowledge, they differ in three respects: (i)~\textit{Source}: Tajik-FactQA is derived from centralized university entrance examinations, whereas Tajik History is derived from school textbooks; (ii)~\textit{Scope}: Tajik-FactQA covers culture, geography, and institutions in addition to history, whereas Tajik History focuses exclusively on historical events and periods; and (iii)~\textit{Difficulty}: Tajik-FactQA targets a post-secondary examination level, whereas Tajik History spans grades 5--11.

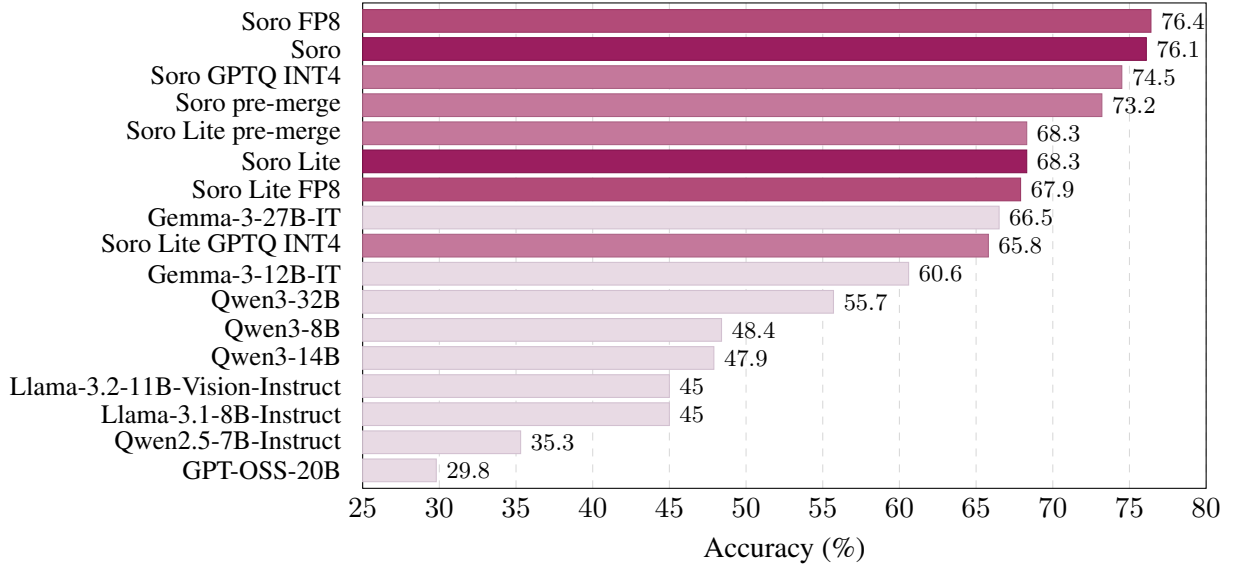
\begin{figure}[H]
    \centering
    \begin{tikzpicture}
    \begin{axis}[
        xbar,
        /pgf/bar shift=0pt,
        width=0.8\textwidth,
        height=8cm,
        xlabel={Accuracy (\%)},
        symbolic y coords={
            GPT-OSS-20B,
            Qwen2.5-7B-Instruct,
            Llama-3.1-8B-Instruct,
            Llama-3.2-11B-Vision-Instruct,
            Qwen3-14B,
            Qwen3-8B,
            Qwen3-32B,
            Gemma-3-12B-IT,
            Soro Lite GPTQ INT4,
            Gemma-3-27B-IT,
            Soro Lite FP8,
            Soro Lite,
            Soro Lite pre-merge,
            Soro pre-merge,
            Soro GPTQ INT4,
            Soro,
            Soro FP8
        },
        ytick=data,
        yticklabel style={font=\small},
        nodes near coords,
        nodes near coords align={horizontal},
        nodes near coords style={font=\footnotesize},
        every node near coord/.append style={
            /pgf/number format/.cd,
            fixed,
            precision=1,
            /tikz/.cd
        },
        xmin=25,
        xmax=80,
        bar width=0.3cm,
        enlarge y limits=0.04,
        axis line style={-},
        tick style={draw=none},
        ymajorgrids=false,
        xmajorgrids=true,
        grid style={dashed, gray!30},
    ]
    
    \addplot[draw=none, fill=none, forget plot] coordinates {
        (0,GPT-OSS-20B) (0,Qwen2.5-7B-Instruct) (0,Llama-3.1-8B-Instruct) (0,Llama-3.2-11B-Vision-Instruct) (0,Qwen3-14B) (0,Qwen3-8B) (0,Qwen3-32B) (0,Gemma-3-12B-IT) (0,Soro Lite GPTQ INT4) (0,Gemma-3-27B-IT) (0,Soro Lite FP8) (0,Soro Lite) (0,Soro Lite pre-merge) (0,Soro pre-merge) (0,Soro GPTQ INT4) (0,Soro) (0,Soro FP8)
    };
    
    \addplot[fill=othermodel, draw=othermodelborder] coordinates {
        (29.8,GPT-OSS-20B)
    };
    
    \addplot[fill=othermodel, draw=othermodelborder] coordinates {
        (35.3,Qwen2.5-7B-Instruct)
        (47.9,Qwen3-14B)
        (48.4,Qwen3-8B)
        (55.7,Qwen3-32B)
    };
    
    \addplot[fill=othermodel, draw=othermodelborder] coordinates {
        (45.0,Llama-3.1-8B-Instruct)
        (45.0,Llama-3.2-11B-Vision-Instruct)
    };
    
    \addplot[fill=othermodel, draw=othermodelborder] coordinates {
        (60.6,Gemma-3-12B-IT)
        (66.5,Gemma-3-27B-IT)
    };
    
    \addplot[fill=sorofamily, draw=sorofamilyborder] coordinates {
        (65.8,Soro Lite GPTQ INT4)
        (74.5,Soro GPTQ INT4)
    };
    
    \addplot[fill=sorofp8, draw=sorofp8border] coordinates {
        (67.9,Soro Lite FP8)
        (76.4,Soro FP8)
    };
    
    \addplot[fill=soromain, draw=soromainborder] coordinates {
        (68.3,Soro Lite)
        (76.1,Soro)
    };
    
    \addplot[fill=sorofamily, draw=sorofamilyborder] coordinates {
        (68.3,Soro Lite pre-merge)
        (73.2,Soro pre-merge)
    };
    
    \end{axis}
    \end{tikzpicture}
    \caption{Tajik Factual Benchmark}
    \label{fig:taj-factual-performance}
\end{figure}

\subsection{Tajik History (high-school level)}
\label{app:tajik-history}

The Tajik History benchmark contains 1{,}400 multiple-choice questions covering the national history curriculum for grades 5--11. Questions derive from officially approved school textbooks and span topics including ancient and medieval Central Asian civilizations, the Samanid dynasty, the formation of modern Tajikistan, Soviet-era history, and post-independence developments. The grade-stratified design enables fine-grained assessment of historical knowledge at varying levels of depth and complexity. Compared to Tajik-FactQA (Appendix~\ref{app:tajik-factual}), this benchmark is narrower in domain (history only) but substantially larger in size and broader in chronological coverage. This dataset is available on
Hugging Face.\footnote{\url{https://huggingface.co/datasets/zehnlab/tajik-history-grades-benchmark}}

\begin{figure}[H]
    \centering
    \begin{tikzpicture}
    \begin{axis}[
        xbar,
        /pgf/bar shift=0pt,
        width=0.8\textwidth,
        height=8cm,
        xlabel={Accuracy (\%)},
        symbolic y coords={
            GPT-OSS-20B,
            Qwen2.5-7B-Instruct,
            Qwen3-8B,
            Llama-3.1-8B-Instruct,
            Llama-3.2-11B-Vision-Instruct,
            Qwen3-14B,
            Gemma-3-12B-IT,
            Qwen3-32B,
            Soro Lite pre-merge,
            Soro Lite GPTQ INT4,
            Soro Lite FP8,
            Gemma-3-27B-IT,
            Soro Lite,
            Soro GPTQ INT4,
            Soro pre-merge,
            Soro,
            Soro FP8
        },
        ytick=data,
        yticklabel style={font=\small},
        nodes near coords,
        nodes near coords align={horizontal},
        nodes near coords style={font=\footnotesize},
        every node near coord/.append style={
            /pgf/number format/.cd,
            fixed,
            precision=1,
            /tikz/.cd
        },
        xmin=25,
        xmax=70,
        bar width=0.3cm,
        enlarge y limits=0.04,
        axis line style={-},
        tick style={draw=none},
        ymajorgrids=false,
        xmajorgrids=true,
        grid style={dashed, gray!30},
    ]
    
    \addplot[draw=none, fill=none, forget plot] coordinates {
        (0,GPT-OSS-20B) (0,Qwen2.5-7B-Instruct) (0,Qwen3-8B) (0,Llama-3.1-8B-Instruct) (0,Llama-3.2-11B-Vision-Instruct) (0,Qwen3-14B) (0,Gemma-3-12B-IT) (0,Qwen3-32B) (0,Soro Lite pre-merge) (0,Soro Lite GPTQ INT4) (0,Soro Lite FP8) (0,Gemma-3-27B-IT) (0,Soro Lite) (0,Soro GPTQ INT4) (0,Soro pre-merge) (0,Soro) (0,Soro FP8)
    };
    
    \addplot[fill=othermodel, draw=othermodelborder] coordinates {
        (34.4,GPT-OSS-20B)
    };
    
    \addplot[fill=othermodel, draw=othermodelborder] coordinates {
        (37.5,Qwen2.5-7B-Instruct)
        (40.5,Qwen3-8B)
        (43.8,Qwen3-14B)
        (46.7,Qwen3-32B)
    };
    
    \addplot[fill=othermodel, draw=othermodelborder] coordinates {
        (42.1,Llama-3.1-8B-Instruct)
        (42.7,Llama-3.2-11B-Vision-Instruct)
    };
    
    \addplot[fill=othermodel, draw=othermodelborder] coordinates {
        (44.2,Gemma-3-12B-IT)
        (49.3,Gemma-3-27B-IT)
    };
    
    \addplot[fill=sorofamily, draw=sorofamilyborder] coordinates {
        (48.7,Soro Lite pre-merge)
        (56.6,Soro pre-merge)
    };
    
    \addplot[fill=sorofamily, draw=sorofamilyborder] coordinates {
        (48.9,Soro Lite GPTQ INT4)
        (55.5,Soro GPTQ INT4)
    };
    
    \addplot[fill=sorofp8, draw=sorofp8border] coordinates {
        (48.9,Soro Lite FP8)
        (57.4,Soro FP8)
    };
    
    \addplot[fill=soromain, draw=soromainborder] coordinates {
        (49.4,Soro Lite)
        (57.2,Soro)
    };
    
    \end{axis}
    \end{tikzpicture}
    \caption{Tajik History high-school level benchmark}
    \label{fig:taj-history-grades-performance}
\end{figure}
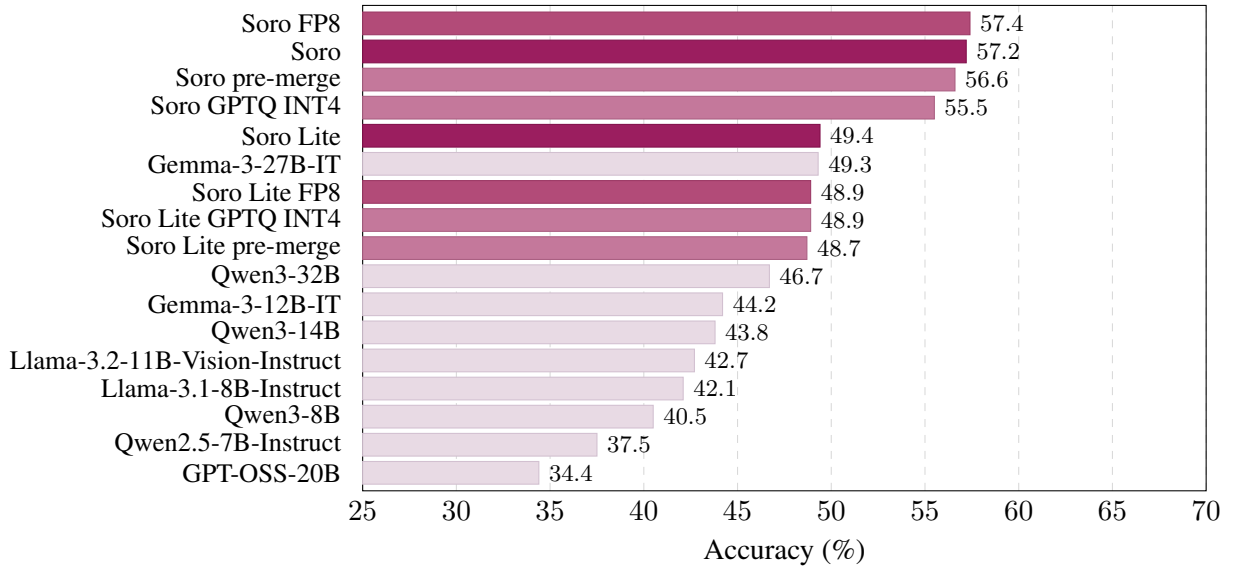

\subsection{Tajik Linguistic Benchmark (TajLib)}
\label{app:tajlib}

TajLib is a 919-question benchmark targeting linguistic competence in Tajik. Questions cover morphological analysis, syntactic structure, lexical semantics, word formation, and orthographic conventions. Unlike the Tajik Language subset within the Curated General Benchmark (Appendix~\ref{app:tajik-curated}), which includes language questions alongside four other domains, TajLib is dedicated entirely to linguistic phenomena and provides substantially deeper coverage of grammatical and structural knowledge.

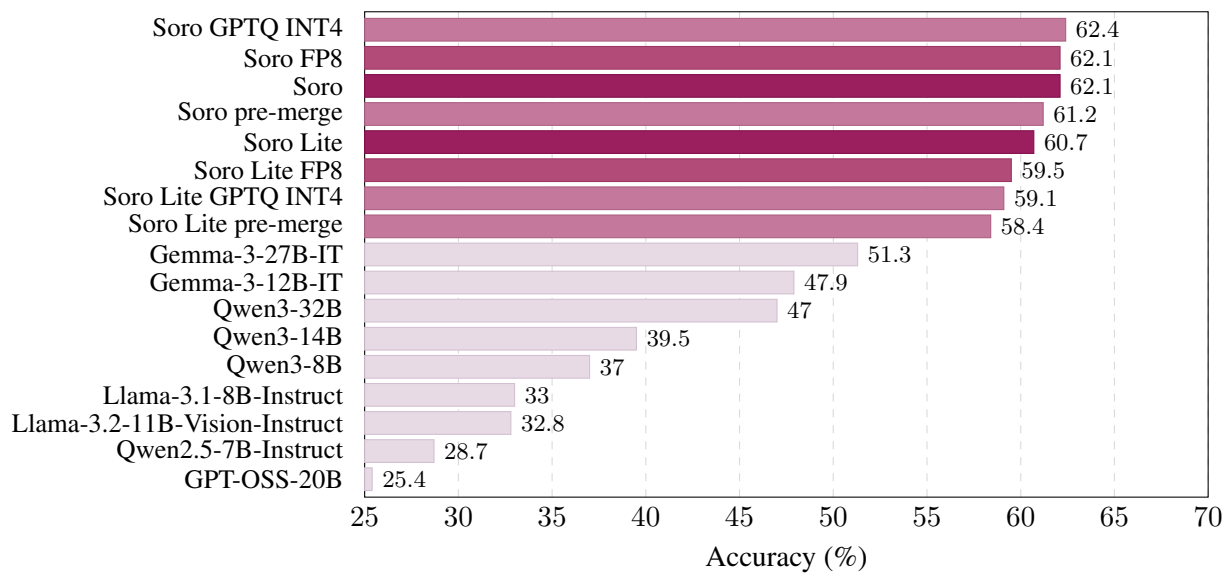
\begin{figure}[H]
    \centering
    \begin{tikzpicture}
    \begin{axis}[
        xbar,
        /pgf/bar shift=0pt,
        width=0.8\textwidth,
        height=8cm,
        xlabel={Accuracy (\%)},
        symbolic y coords={
            GPT-OSS-20B,
            Qwen2.5-7B-Instruct,
            Llama-3.2-11B-Vision-Instruct,
            Llama-3.1-8B-Instruct,
            Qwen3-8B,
            Qwen3-14B,
            Qwen3-32B,
            Gemma-3-12B-IT,
            Gemma-3-27B-IT,
            Soro Lite pre-merge,
            Soro Lite GPTQ INT4,
            Soro Lite FP8,
            Soro Lite,
            Soro pre-merge,
            Soro,
            Soro FP8,
            Soro GPTQ INT4
        },
        ytick=data,
        yticklabel style={font=\small},
        nodes near coords,
        nodes near coords align={horizontal},
        nodes near coords style={font=\footnotesize},
        every node near coord/.append style={
            /pgf/number format/.cd,
            fixed,
            precision=1,
            /tikz/.cd
        },
        xmin=25,
        xmax=70,
        bar width=0.3cm,
        enlarge y limits=0.04,
        axis line style={-},
        tick style={draw=none},
        ymajorgrids=false,
        xmajorgrids=true,
        grid style={dashed, gray!30},
    ]
    
    \addplot[draw=none, fill=none, forget plot] coordinates {
        (0,GPT-OSS-20B) (0,Qwen2.5-7B-Instruct) (0,Llama-3.2-11B-Vision-Instruct) (0,Llama-3.1-8B-Instruct) (0,Qwen3-8B) (0,Qwen3-14B) (0,Qwen3-32B) (0,Gemma-3-12B-IT) (0,Gemma-3-27B-IT) (0,Soro Lite pre-merge) (0,Soro Lite GPTQ INT4) (0,Soro Lite FP8) (0,Soro Lite) (0,Soro pre-merge) (0,Soro) (0,Soro FP8) (0,Soro GPTQ INT4)
    };
    
    \addplot[fill=othermodel, draw=othermodelborder] coordinates {
        (25.4,GPT-OSS-20B)
    };
    
    \addplot[fill=othermodel, draw=othermodelborder] coordinates {
        (28.7,Qwen2.5-7B-Instruct)
        (37.0,Qwen3-8B)
        (39.5,Qwen3-14B)
        (47.0,Qwen3-32B)
    };
    
    \addplot[fill=othermodel, draw=othermodelborder] coordinates {
        (32.8,Llama-3.2-11B-Vision-Instruct)
        (33.0,Llama-3.1-8B-Instruct)
    };
    
    \addplot[fill=othermodel, draw=othermodelborder] coordinates {
        (47.9,Gemma-3-12B-IT)
        (51.3,Gemma-3-27B-IT)
    };
    
    \addplot[fill=sorofamily, draw=sorofamilyborder] coordinates {
        (58.4,Soro Lite pre-merge)
        (61.2,Soro pre-merge)
    };
    
    \addplot[fill=sorofamily, draw=sorofamilyborder] coordinates {
        (59.1,Soro Lite GPTQ INT4)
        (62.4,Soro GPTQ INT4)
    };
    
    \addplot[fill=sorofp8, draw=sorofp8border] coordinates {
        (59.5,Soro Lite FP8)
        (62.1,Soro FP8)
    };
    
    \addplot[fill=soromain, draw=soromainborder] coordinates {
        (60.7,Soro Lite)
        (62.1,Soro)
    };
    
    \end{axis}
    \end{tikzpicture}
    \caption{Tajik Linguistic Benchmark (TajLib)}
    \label{fig:taj-lib-performance}
\end{figure}

This focused design makes TajLib especially diagnostic: strong performance indicates that a model has acquired robust internal representations of Tajik grammar and vocabulary, rather than relying on surface-level lexical patterns. TajLib exhibits the largest performance gap between Soro and baseline models among all benchmarks in our suite, exceeding 10 percentage points (see Figure~\ref{fig:taj-lib-performance}), which points to the pronounced gains that continual pretraining on Tajik text yields in core linguistic competence. This dataset is available on
Hugging Face.\footnote{\url{https://huggingface.co/datasets/zehnlab/TajLib-language-bench}}

\subsection{Tajik Literature (high-school level)}
\label{app:tajik-literature}

The Tajik Literature benchmark contains 698 multiple-choice questions covering the national literature curriculum for grades 5--11. Questions derive from officially approved school textbooks and span classical Tajik poetry, prominent authors and poets of the Persian--Tajik literary tradition, narrative prose, literary devices, and contemporary Tajik literature. The grade-stratified design enables assessment of literary knowledge at varying levels of depth and complexity, from foundational concepts introduced in grade 5 to more advanced critical analysis required in grades 10--11. Compared to the Tajik Curated General Benchmark (Appendix~\ref{app:tajik-curated}), which includes a literature subset alongside other domains, this benchmark focuses exclusively on literary knowledge and provides substantially deeper coverage across the full secondary school curriculum. This
dataset is available on Hugging Face.\footnote{\url{https://huggingface.co/datasets/zehnlab/tajik-literature-grades-benchmark}}

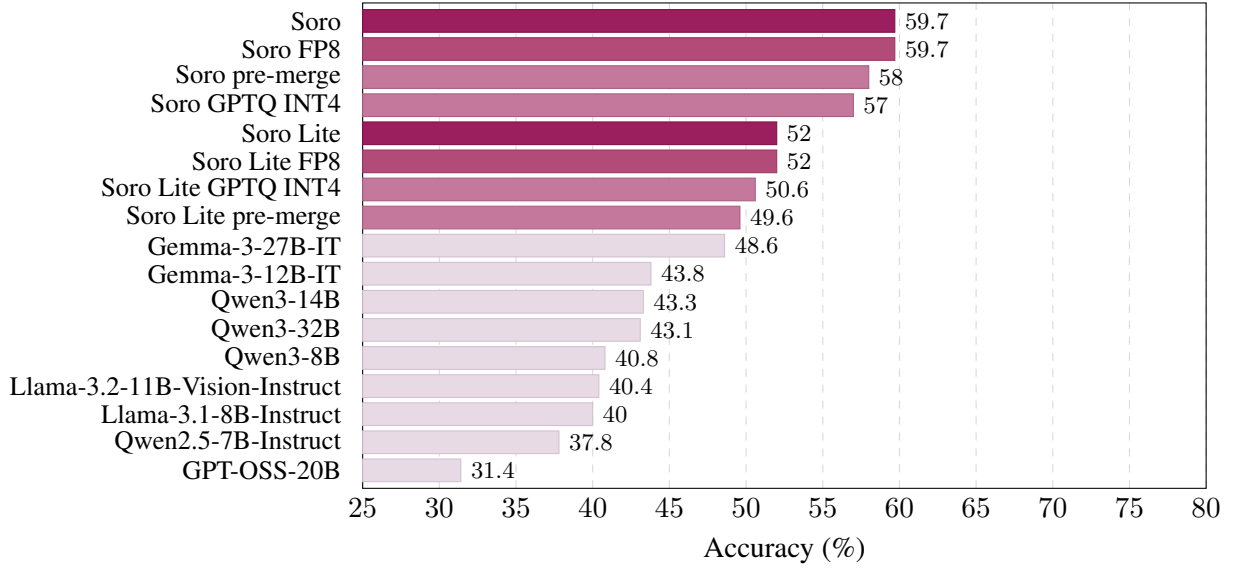
\begin{figure}[H]
    \centering
    \begin{tikzpicture}
    \begin{axis}[
        xbar,
        /pgf/bar shift=0pt,
        width=0.8\textwidth,
        height=8cm,
        xlabel={Accuracy (\%)},
        symbolic y coords={
            GPT-OSS-20B,
            Qwen2.5-7B-Instruct,
            Llama-3.1-8B-Instruct,
            Llama-3.2-11B-Vision-Instruct,
            Qwen3-8B,
            Qwen3-32B,
            Qwen3-14B,
            Gemma-3-12B-IT,
            Gemma-3-27B-IT,
            Soro Lite pre-merge,
            Soro Lite GPTQ INT4,
            Soro Lite FP8,
            Soro Lite,
            Soro GPTQ INT4,
            Soro pre-merge,
            Soro FP8,
            Soro
        },
        ytick=data,
        yticklabel style={font=\small},
        nodes near coords,
        nodes near coords align={horizontal},
        nodes near coords style={font=\footnotesize},
        every node near coord/.append style={
            /pgf/number format/.cd,
            fixed,
            precision=1,
            /tikz/.cd
        },
        xmin=25,
        xmax=80,
        bar width=0.3cm,
        enlarge y limits=0.04,
        axis line style={-},
        tick style={draw=none},
        ymajorgrids=false,
        xmajorgrids=true,
        grid style={dashed, gray!30},
    ]
    
    \addplot[draw=none, fill=none, forget plot] coordinates {
        (0,GPT-OSS-20B) (0,Qwen2.5-7B-Instruct) (0,Llama-3.1-8B-Instruct) (0,Llama-3.2-11B-Vision-Instruct) (0,Qwen3-8B) (0,Qwen3-32B) (0,Qwen3-14B) (0,Gemma-3-12B-IT) (0,Gemma-3-27B-IT) (0,Soro Lite pre-merge) (0,Soro Lite GPTQ INT4) (0,Soro Lite FP8) (0,Soro Lite) (0,Soro GPTQ INT4) (0,Soro pre-merge) (0,Soro FP8) (0,Soro)
    };
    
    \addplot[fill=othermodel, draw=othermodelborder] coordinates {
        (31.4,GPT-OSS-20B)
    };
    
    \addplot[fill=othermodel, draw=othermodelborder] coordinates {
        (37.8,Qwen2.5-7B-Instruct)
        (40.8,Qwen3-8B)
        (43.1,Qwen3-32B)
        (43.3,Qwen3-14B)
    };
    
    \addplot[fill=othermodel, draw=othermodelborder] coordinates {
        (40.0,Llama-3.1-8B-Instruct)
        (40.4,Llama-3.2-11B-Vision-Instruct)
    };
    
    \addplot[fill=othermodel, draw=othermodelborder] coordinates {
        (43.8,Gemma-3-12B-IT)
        (48.6,Gemma-3-27B-IT)
    };
    
    \addplot[fill=sorofamily, draw=sorofamilyborder] coordinates {
        (49.6,Soro Lite pre-merge)
        (58.0,Soro pre-merge)
    };
    
    \addplot[fill=sorofamily, draw=sorofamilyborder] coordinates {
        (50.6,Soro Lite GPTQ INT4)
        (57.0,Soro GPTQ INT4)
    };
    
    \addplot[fill=sorofp8, draw=sorofp8border] coordinates {
        (52.0,Soro Lite FP8)
        (59.7,Soro FP8)
    };
    
    \addplot[fill=soromain, draw=soromainborder] coordinates {
        (52.0,Soro Lite)
        (59.7,Soro)
    };
    
    \end{axis}
    \end{tikzpicture}
    \caption{Tajik Literature Benchmark}
    \label{fig:taj-literature-performance}
\end{figure}

\subsection{Tajik MMLU}
\label{app:tajik-mmlu}
Tajik MMLU is a Tajik-language translation of the original English MMLU benchmark, covering the same 57 subject categories and preserving the original question structure and answer options. Its role is complementary to English MMLU: while the latter measures retention of general knowledge, Tajik MMLU measures whether the model can access and apply the \textit{same} knowledge when prompted in Tajik. Gaps between the two scores isolate the effect of language on reasoning and knowledge retrieval, independent of domain coverage. The benchmark contains the same number of questions as the English variant.  This
dataset is available on Hugging Face.\footnote{\url{https://huggingface.co/datasets/zehnlab/tajik-mmlu}}

\begin{figure}[H]
    \centering
    \begin{tikzpicture}
    \begin{axis}[
        xbar,
        /pgf/bar shift=0pt,
        width=0.8\textwidth,
        height=8cm,
        xlabel={Accuracy (\%)},
        symbolic y coords={
            GPT-OSS-20B,
            Qwen2.5-7B-Instruct,
            Llama-3.2-11B-Vision-Instruct,
            Llama-3.1-8B-Instruct,
            Qwen3-8B,
            Qwen3-14B,
            Soro Lite pre-merge,
            Gemma-3-12B-IT,
            Soro Lite GPTQ INT4,
            Qwen3-32B,
            Soro Lite FP8,
            Soro Lite,
            Soro pre-merge,
            Soro GPTQ INT4,
            Gemma-3-27B-IT,
            Soro FP8,
            Soro
        },
        ytick=data,
        yticklabel style={font=\small},
        nodes near coords,
        nodes near coords align={horizontal},
        nodes near coords style={font=\footnotesize},
        every node near coord/.append style={
            /pgf/number format/.cd,
            fixed,
            precision=1,
            /tikz/.cd
        },
        xmin=25,
        xmax=80,
        bar width=0.3cm,
        enlarge y limits=0.04,
        axis line style={-},
        tick style={draw=none},
        ymajorgrids=false,
        xmajorgrids=true,
        grid style={dashed, gray!30},
    ]
    
    \addplot[draw=none, fill=none, forget plot] coordinates {
        (0,GPT-OSS-20B) (0,Qwen2.5-7B-Instruct) (0,Llama-3.2-11B-Vision-Instruct) (0,Llama-3.1-8B-Instruct) (0,Qwen3-8B) (0,Qwen3-14B) (0,Soro Lite pre-merge) (0,Gemma-3-12B-IT) (0,Soro Lite GPTQ INT4) (0,Qwen3-32B) (0,Soro Lite FP8) (0,Soro Lite) (0,Soro pre-merge) (0,Soro GPTQ INT4) (0,Gemma-3-27B-IT) (0,Soro FP8) (0,Soro)
    };
    
    \addplot[fill=othermodel, draw=othermodelborder] coordinates {
        (31.8,GPT-OSS-20B)
    };
    
    \addplot[fill=othermodel, draw=othermodelborder] coordinates {
        (35.5,Qwen2.5-7B-Instruct)
        (47.7,Qwen3-8B)
        (51.3,Qwen3-14B)
        (61.4,Qwen3-32B)
    };
    
    \addplot[fill=othermodel, draw=othermodelborder] coordinates {
        (36.9,Llama-3.2-11B-Vision-Instruct)
        (37.5,Llama-3.1-8B-Instruct)
    };
    
    \addplot[fill=sorofamily, draw=sorofamilyborder] coordinates {
        (59.3,Soro Lite pre-merge)
        (67.1,Soro pre-merge)
    };
    
    \addplot[fill=othermodel, draw=othermodelborder] coordinates {
        (59.7,Gemma-3-12B-IT)
        (67.6,Gemma-3-27B-IT)
    };
    
    \addplot[fill=sorofamily, draw=sorofamilyborder] coordinates {
        (60.9,Soro Lite GPTQ INT4)
        (67.1,Soro GPTQ INT4)
    };
    
    \addplot[fill=sorofp8, draw=sorofp8border] coordinates {
        (62.4,Soro Lite FP8)
        (69.4,Soro FP8)
    };
    
    \addplot[fill=soromain, draw=soromainborder] coordinates {
        (62.8,Soro Lite)
        (69.5,Soro)
    };
    
    \end{axis}
    \end{tikzpicture}
    \caption{Tajik MMLU Benchmark}
    \label{fig:taj-mmlu-performance}
\end{figure}
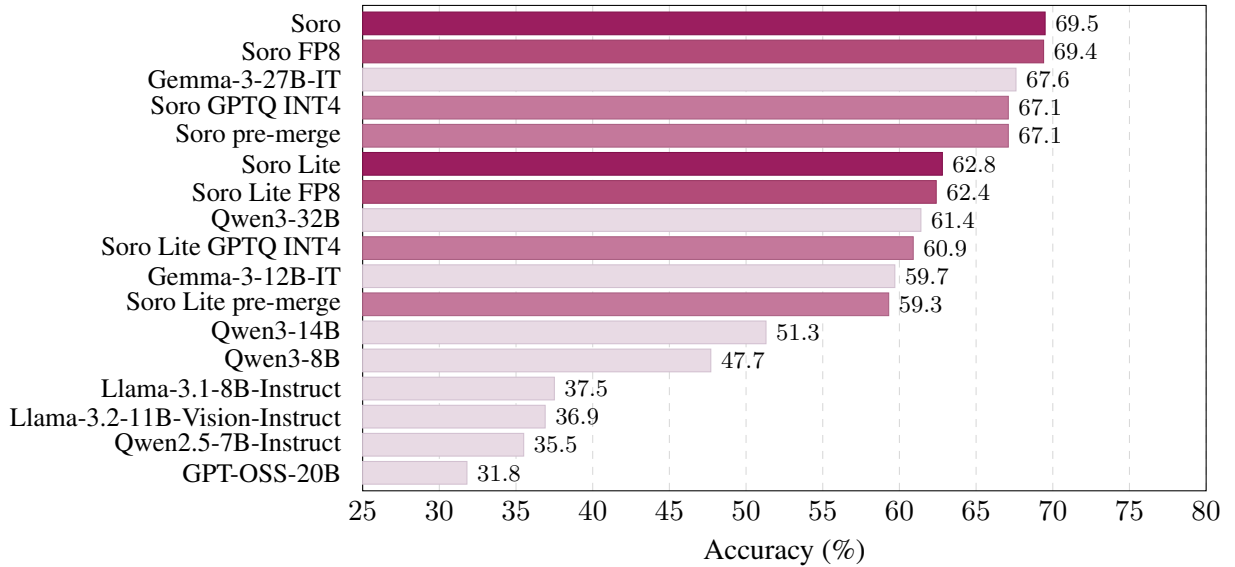

\section{Evaluation Implementation Details}
\label{sec:eval_implementation}

This section describes the technical details of our evaluation framework for multiple-choice benchmarks (MMLU, Tajik-FactQA, Tajik MMLU, etc.). We use two evaluation formats.

\subsection{Raw Format}

\begin{figure}[H]
    \centering
    \includegraphics[width=0.8\textwidth]{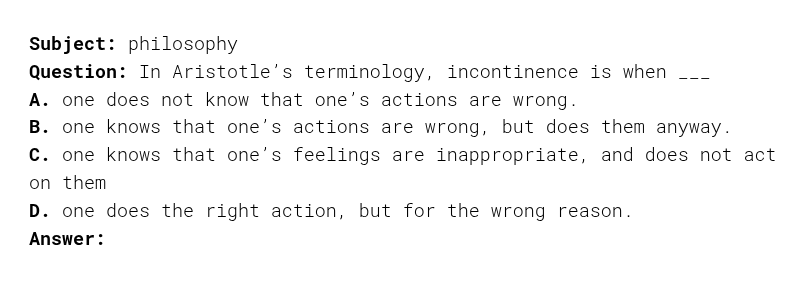}
    \caption{Raw format prompt structure for English evaluation.}
    \label{fig:raw_en_prompt}
\end{figure}

\begin{figure}[H]
    \centering
    \includegraphics[width=0.8\textwidth]{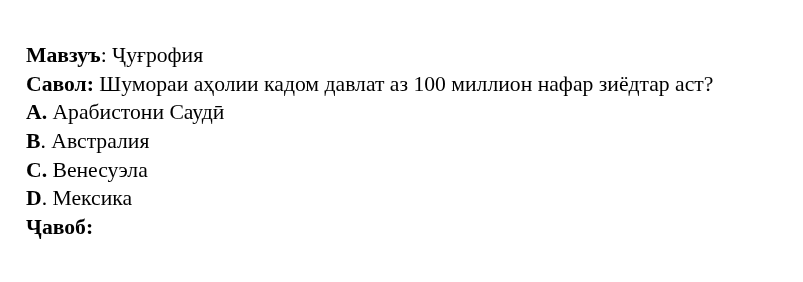}
    \caption{Raw format prompt structure for Tajik evaluation.}
    \label{fig:raw_taj_prompt}
\end{figure}

The raw format is applicable only to models with open weights, as it requires access to token probabilities. In this format, the model receives a plain-text prompt without any chat template. The prompt structure for English and Tajik evaluations is shown in Figures~\ref{fig:raw_en_prompt} and~\ref{fig:raw_taj_prompt}, respectively. For Tajik prompts, the format uses Tajik labels while retaining English letters (A, B, C, D) for answer choices.

\subsection{Chat Format}

\begin{figure}[H]
    \centering
    \includegraphics[width=0.8\textwidth]{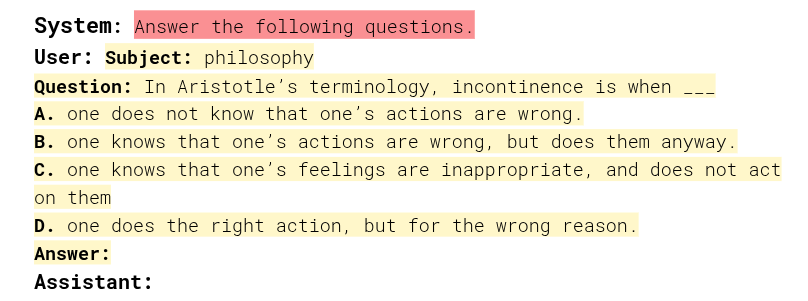}
    \caption{Chat format prompt structure for English evaluation.}
    \label{fig:chat_en_prompt}
\end{figure}

\begin{figure}[H]
    \centering
    \includegraphics[width=0.8\textwidth]{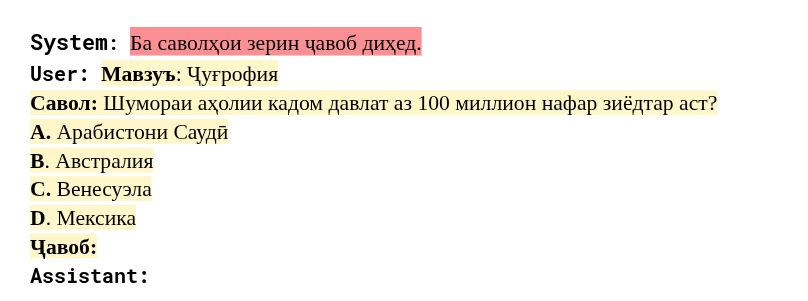}
    \caption{Chat format prompt structure for Tajik evaluation.}
    \label{fig:chat_taj_prompt}
\end{figure}

In chat format, the same content is wrapped in a conversational structure with system and user messages and processed through the tokenizer's chat template. The message structure for English and Tajik evaluations is illustrated in Figures~\ref{fig:chat_en_prompt} and~\ref{fig:chat_taj_prompt}, respectively.

\subsection{Answer Probability Extraction}

For both formats, the evaluation extracts logits from the last token position and computes softmax probabilities over the answer tokens. Multiple token variations for each answer letter are checked (with different prefixes: empty string, space, or newline) to handle tokenization differences across models. The maximum probability across these variations is selected for each answer choice and normalized to form a proper distribution over the four options. The predicted answer is the choice with the highest probability.

\section{Quantization Details}
\label{sec:quantization_details}
The merged Soro instruction-tuned variants were selected for quantization as they achieved the best overall performance across our evaluation suite. We use the \texttt{llmcompressor}~\cite{llmcompressor2024} Python library throughout.

\subsection{Dynamic FP8}

We apply dynamic FP8 quantization~\citep{kuzmin2024fp8quantizationpowerexponent} by converting both weights and activations from BF16 to 8-bit floating-point format at inference time. Quantization targets all \texttt{Linear} layers across the transformer blocks, while the language modeling head (\texttt{lm\_head}) is deliberately excluded to preserve output distribution quality. The scheme uses a calibration-free one-shot approach: weights are quantized directly from the merged checkpoint without any additional training steps. This reduces memory consumption by approximately 50\% (from 24\,GB to 13\,GB for Soro Lite, and from 54\,GB to 29\,GB for Soro) with negligible accuracy loss, as confirmed by our Tajik benchmark evaluations in Table~\ref{tab:quant_model_comparison}.

\subsection{GPTQ INT4}

GPTQ quantization~\citep{frantar2023gptqaccurateposttrainingquantization} compresses model weights to 4-bit integers using a calibration-based second-order optimization procedure. Unlike FP8, GPTQ requires a small representative dataset to compute activation statistics used to minimize quantization error. We use 128 samples drawn from Tajik Wikipedia (the \texttt{finewiki} dataset, Tajik subset~\citep{penedo2025finewiki}), ensuring that the calibration distribution is linguistically aligned with the target deployment context rather than generic English data.

Quantization is applied group-wise with a group size of 128, using symmetric integer quantization across all \texttt{Linear} layers. Several components are explicitly excluded: the language modeling head (\texttt{lm\_head}), the token embedding layer (\texttt{embed\_tokens}), the vision encoder (\texttt{vision\_tower}), and the multimodal projector (\texttt{multi\_modal\_projector}). This configuration achieves a $4\times$ memory reduction (from 24\,GB to 6.6\,GB for Soro Lite, and from 54\,GB to 13.5\,GB for Soro) while retaining the majority of Tajik benchmark performance gains over the Gemma~3 baselines, as shown in Table~\ref{tab:quant_model_comparison}.

\section{Sample Interactions}
\label{app:sample-interactions}

This section presents representative interactions with Soro (27B) to illustrate its conversational capabilities across domains relevant to the educational deployment context. Each example shows a screenshot of the user prompt and model response in Tajik, followed by an English translation. All outputs were generated using greedy decoding (temperature~$= 0$) from the final merged Soro checkpoint; no content has been edited.

\definecolor{soroboxbg}{RGB}{240, 215, 228}
\definecolor{soroboxframe}{RGB}{178, 75, 120}

\newtcolorbox{translationbox}{
    colback=white,
    colframe=gray!50,
    arc=2pt,
    boxrule=0.4pt,
    left=10pt, right=10pt, top=6pt, bottom=6pt,
    fonttitle=\bfseries\small\itshape,
    title={English Translation},
}


\subsection*{Example 1: Tajik History}
\label{app:sample-history}

\begin{figure}[H]
    \centering
    \fcolorbox{gray!40}{gray!8}{\includegraphics[width=0.9\textwidth]{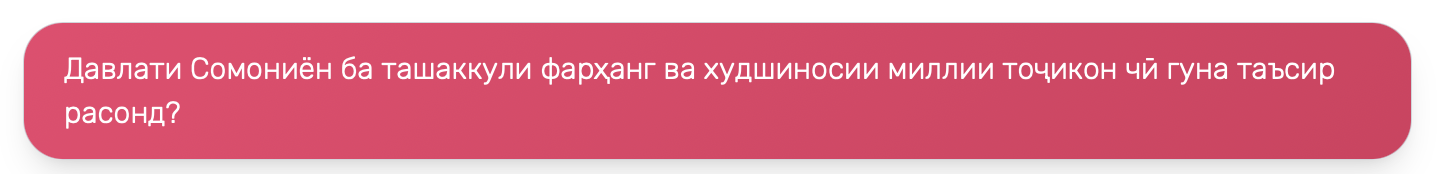}}
    \vspace{2pt}
    \fcolorbox{soroboxframe}{soroboxbg}{\includegraphics[width=0.9\textwidth]{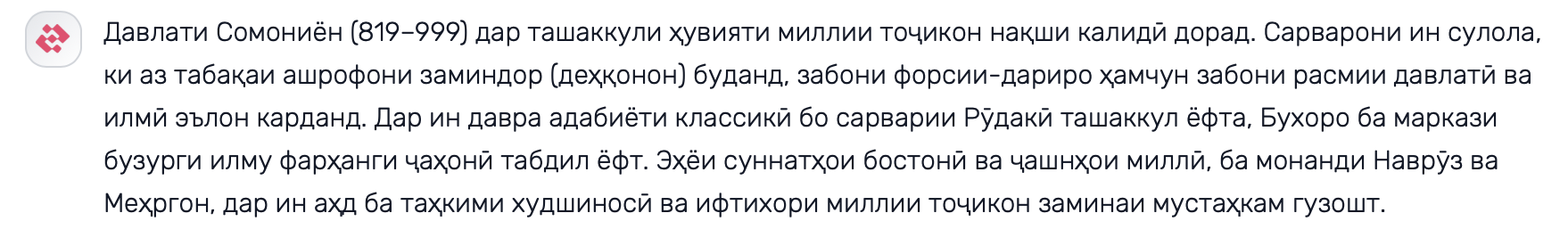}}
\end{figure}

\begin{translationbox}
\textbf{User:} How did the Samanid state influence the formation of Tajik culture and national identity?

\medskip
\textbf{Soro:} The Samanid State (819--999) played a key role in the formation of Tajik national identity. The rulers of this dynasty, who hailed from the landowning aristocracy (dehkans), declared Persian-Dari as the official state and scientific language. During this period, classical literature flourished under the leadership of Rudaki, and Bukhara transformed into a major global center for science and culture. The revival of ancient traditions and national festivals, such as Navruz and Mehrgon, established a solid foundation for the strengthening of Tajik self-awareness and national pride.
\end{translationbox}


\subsection*{Example 2: School-Level Science}
\label{app:sample-science}

\begin{figure}[H]
    \centering
    \fcolorbox{gray!40}{gray!8}{\includegraphics[width=0.9\textwidth]{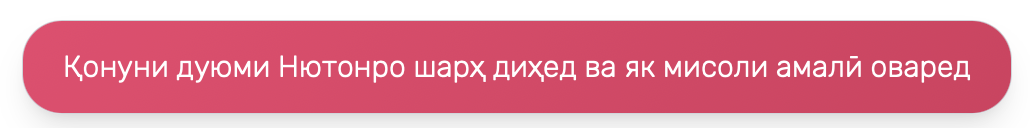}}
    \vspace{2pt}
    \fcolorbox{soroboxframe}{soroboxbg}{\includegraphics[width=0.9\textwidth]{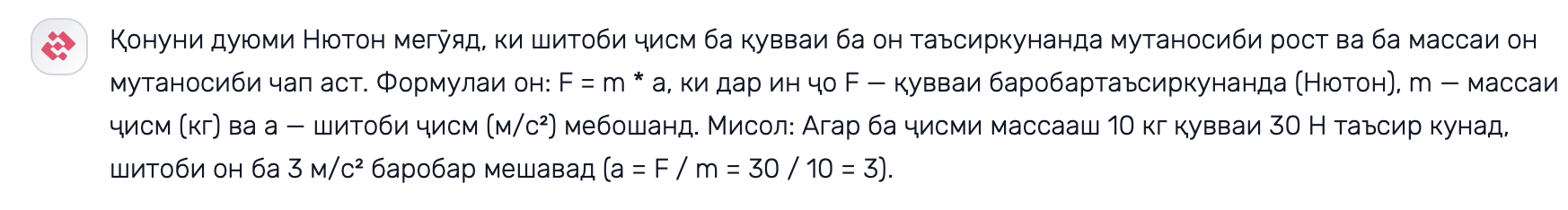}}
\end{figure}

\begin{translationbox}
\textbf{User:} Explain Newton's second law and provide a practical example.

\medskip
\textbf{Soro:} Newton's Second Law states that the acceleration of an object is directly proportional to the net force acting upon it and inversely proportional to its mass. Its formula is $F = m \cdot a$, where $F$ is the resultant force (Newtons), $m$ is the mass of the object (kg), and $a$ is the acceleration (m/s\textsuperscript{2}). For example, if a force of 30~N is applied to an object with a mass of 10~kg, its acceleration will be 3~m/s\textsuperscript{2} ($a = F/m = 30/10 = 3$).
\end{translationbox}


\subsection*{Example 3: Tajik Literature}
\label{app:sample-literature}

\begin{figure}[H]
    \centering
    \fcolorbox{gray!40}{gray!8}{\includegraphics[width=0.9\textwidth]{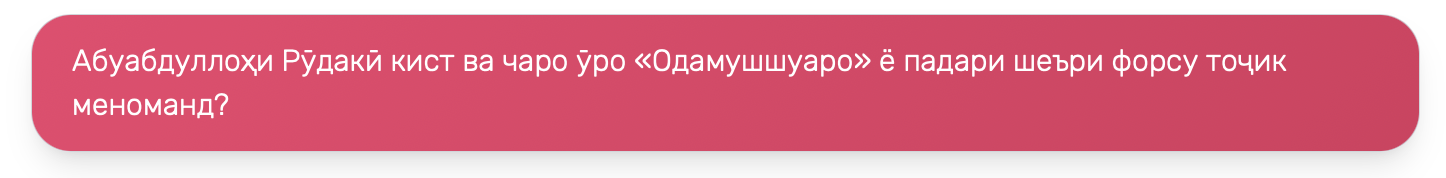}}
    \vspace{2pt}
    \fcolorbox{soroboxframe}{soroboxbg}{\includegraphics[width=0.9\textwidth]{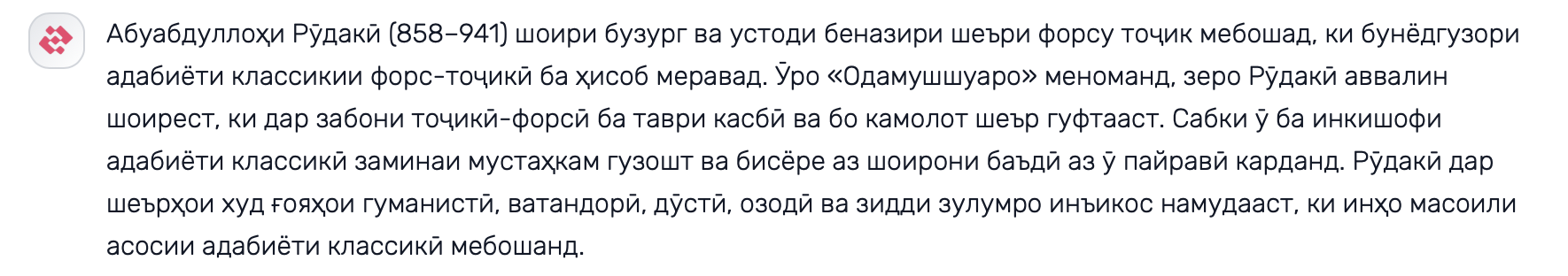}}
\end{figure}

\begin{translationbox}
\textbf{User:} Who is Abu Abdullah Rudaki and why is he referred to as ``Adamush-shu'aro'' (the forefather of poets) or the father of Persian-Tajik poetry?

\medskip
\textbf{Soro:} Abu Abdullah Rudaki (858--941) was a great poet and an unparalleled master of Persian-Tajik verse, regarded as the founder of classical Persian-Tajik literature. He is called ``Adamush-shu'aro'' (the Forefather of Poets) because he was the first to compose poetry in the Tajik-Persian language with professional mastery and perfection. His style laid a strong foundation for the development of classical literature, and many subsequent poets followed his lead. In his poems, Rudaki reflected humanistic ideas, patriotism, friendship, freedom, and opposition to oppression, themes that became the core of classical literature.
\end{translationbox}


\subsection*{Example 4: Geography and Culture of Tajikistan}
\label{app:sample-culture}

\begin{figure}[H]
    \centering
    \fcolorbox{gray!40}{gray!8}{\includegraphics[width=0.9\textwidth]{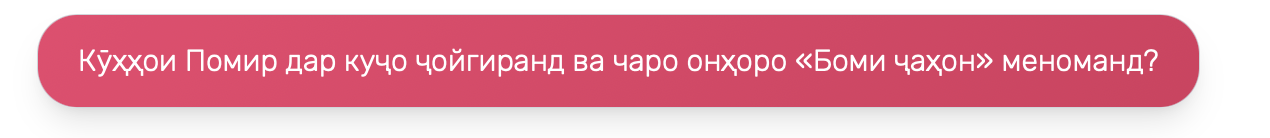}}
    \vspace{2pt}
    \fcolorbox{soroboxframe}{soroboxbg}{\includegraphics[width=0.9\textwidth]{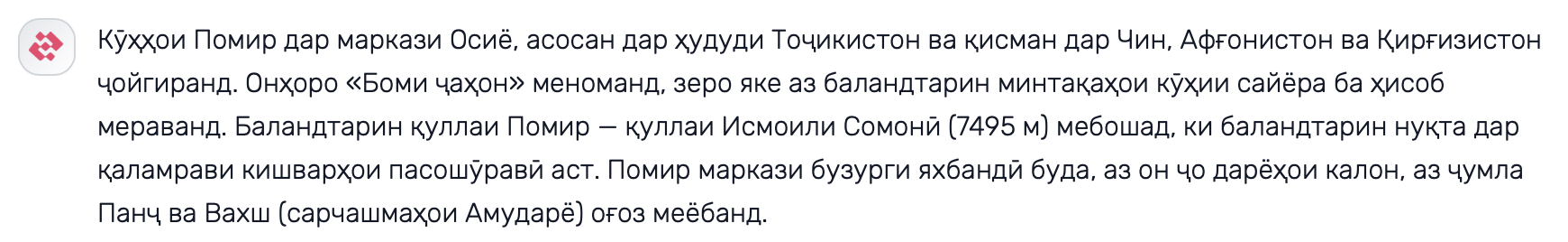}}
\end{figure}

\begin{translationbox}
\textbf{User:} Where are the Pamir Mountains located and why are they known as the ``Roof of the World''?

\medskip
\textbf{Soro:} The Pamir Mountains are located in Central Asia, primarily within the territory of Tajikistan and partially in China, Afghanistan, and Kyrgyzstan. They are known as the ``Roof of the World'' because they are among the highest mountain regions on the planet. The highest peak of the Pamirs is Ismoil Somoni Peak (7495~m), which is the highest point in the former Soviet Union. The Pamirs are a major center of glaciation, serving as the source of large rivers including the Panj and Vakhsh (the headwaters of the Amu Darya).
\end{translationbox}


\subsection*{Example 5: Tajik Language and Grammar}
\label{app:sample-language}

\begin{figure}[H]
    \centering
    \fcolorbox{gray!40}{gray!8}{\includegraphics[width=0.9\textwidth]{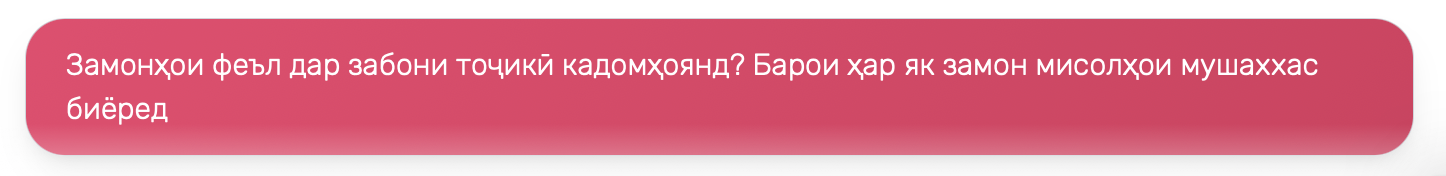}}
    \vspace{2pt}
    \fcolorbox{soroboxframe}{soroboxbg}{\includegraphics[width=0.9\textwidth]{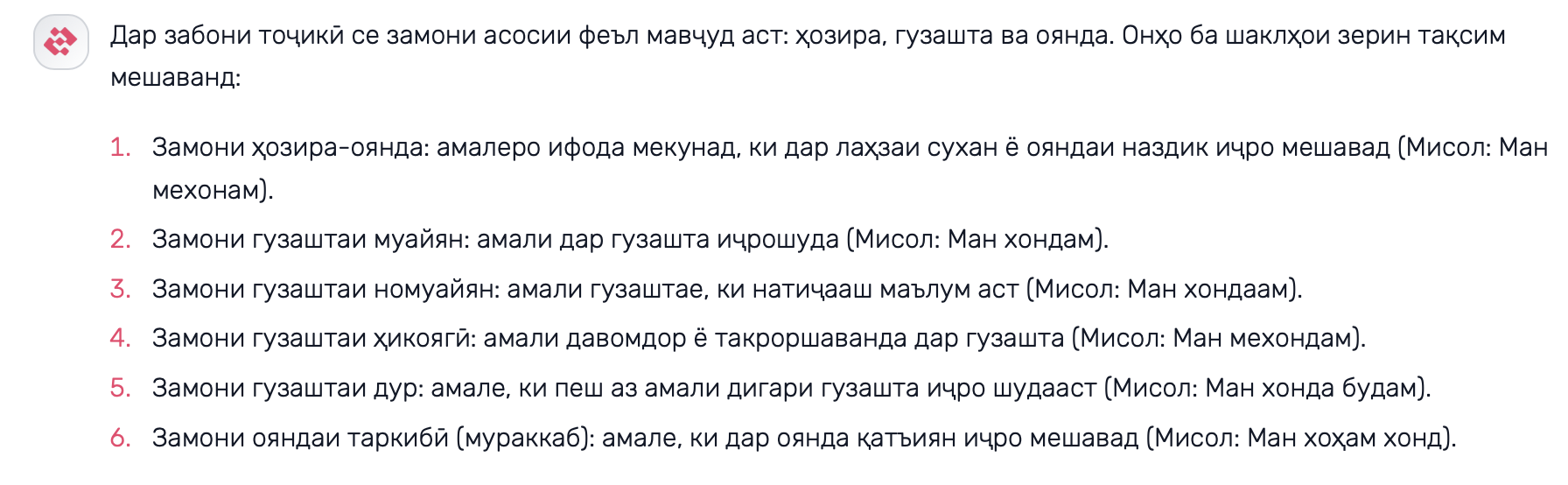}}
\end{figure}

\begin{translationbox}
\textbf{User:} What are the verb tenses in the Tajik language? Provide specific examples for each tense.

\medskip
\textbf{Soro:} In the Tajik language, there are three main verb tenses: present, past, and future. They are categorized into the following forms:

1.~Present-Future Tense: expresses an action occurring at the moment of speech or in the near future (Example: \textit{Man mekhonam} - I read / I will read).

2.~Past Definite Tense: an action completed in the past (Example: \textit{Man khondam} - I read).

3.~Past Indefinite Tense: a past action with a known result or relevance to the present (Example: \textit{Man khondaam} - I have read).

4.~Past Continuous/Narrative Tense: a continuous or repetitive action in the past (Example: \textit{Man mekhondam} - I used to read / I was reading).

5.~Past Perfect (Remote Past): an action completed before another past action (Example: \textit{Man khonda budam} - I had read).

6.~Compound Future Tense: an action that will strictly take place in the future (Example: \textit{Man khoham khond} - I shall read).
\end{translationbox}

 
\subsection*{Example 6: Multi-Turn Conversation --- Lake Iskanderkul}
\label{app:sample-multiturn-iskanderkul}
 
\begin{figure}[H]
    \centering
    \fcolorbox{gray!40}{gray!8}{\includegraphics[width=0.9\textwidth]{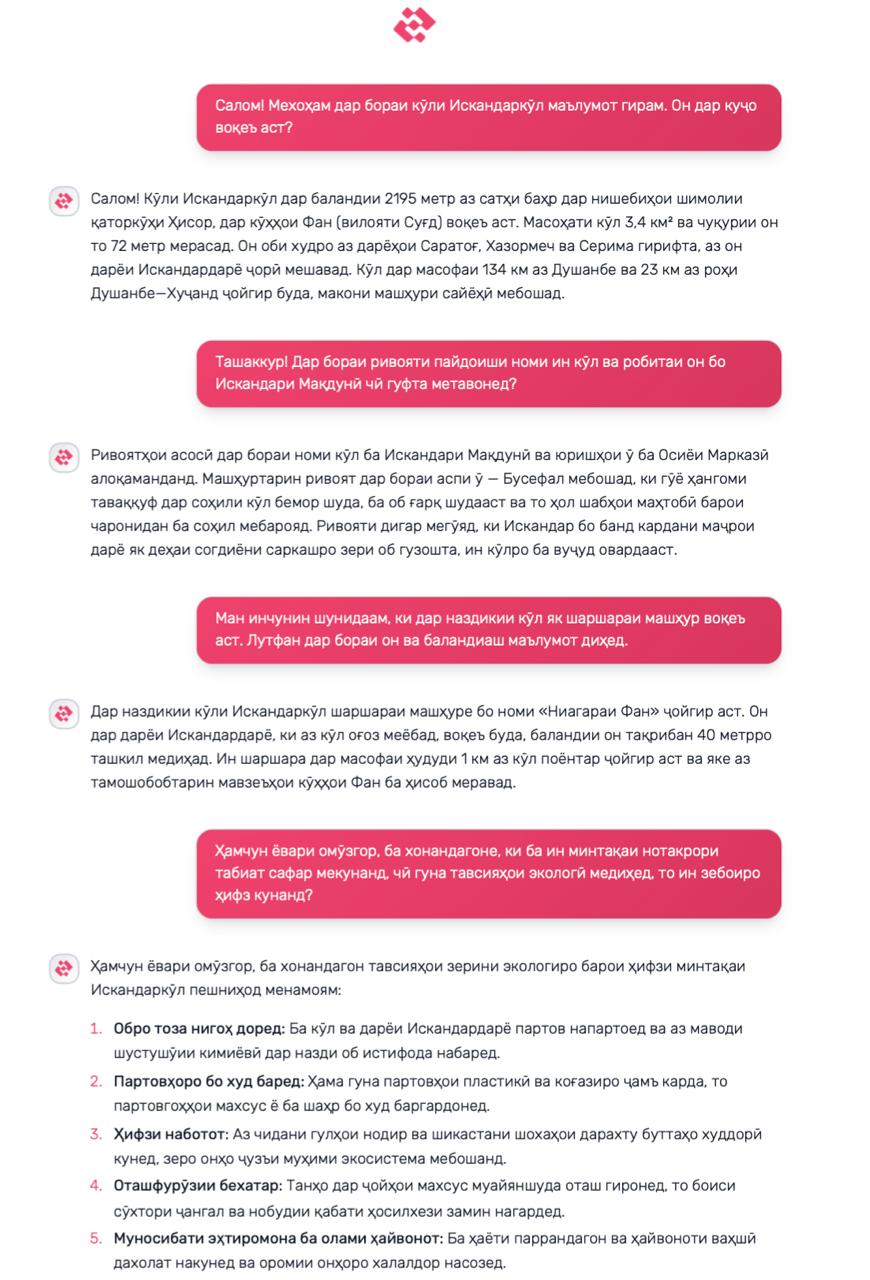}}
\end{figure}
 
\begin{translationbox}
\textbf{User (Turn 1):} Hello! I would like to get information about Lake Iskanderkul. Where is it located?
 
\medskip
\textbf{Soro (Turn 1):} Hello! Lake Iskanderkul is located at an altitude of 2195 meters above sea level on the northern slopes of the Gissar Range, in the Fann Mountains (Sughd region). The lake's area is 3.4~km\textsuperscript{2} and its depth reaches up to 72 meters. It is fed by the Saratogh, Hazormech, and Serima rivers, and the Iskandardarya River flows out of it. The lake is located at a distance of 134~km from Dushanbe and 23~km from the Dushanbe--Khujand highway, making it a famous tourist destination.
 
\medskip
\textbf{User (Turn 2):} Thank you! What can you tell me about the legend of the name's origin and its connection to Alexander the Great?
 
\medskip
\textbf{Soro (Turn 2):} The main legends regarding the lake's name are associated with Alexander the Great and his campaigns in Central Asia. The most famous legend concerns his horse, Bucephalus, which reportedly fell ill during a stop on the lakeshore, drowned in the water, and to this day comes out to graze on the shore on moonlit nights. Another legend states that Alexander created this lake by damming the river to submerge a village of rebellious Sogdians.
 
\medskip
\textbf{User (Turn 3):} I also heard that there is a famous waterfall nearby. Please provide information about it and its height.
 
\medskip
\textbf{Soro (Turn 3):} Near Lake Iskanderkul, there is a famous waterfall known as the ``Fann Niagara.'' It is located on the Iskandardarya River, which originates from the lake, and its height is approximately 40 meters. This waterfall is situated about 1~km downstream from the lake and is considered one of the most spectacular sights in the Fann Mountains.
 
\medskip
\textbf{User (Turn 4):} As a teacher's assistant, what ecological recommendations would you give to students traveling to this unique natural region to preserve its beauty?
 
\medskip
\textbf{Soro (Turn 4):} As a teacher's assistant, I propose the following ecological recommendations for the preservation of the Iskanderkul region:
 
1.~Keep the water clean: Do not throw trash into the lake or the Iskandardarya river and avoid using chemical detergents near the water.
 
2.~Take your trash with you: Collect all plastic and paper waste to return it to special trash collection points or to the city.
 
3.~Protect the flora: Avoid picking rare flowers or breaking tree branches, as they are vital components of the ecosystem.
 
4.~Fire safety: Light fires only in designated areas to prevent forest fires and avoid damaging the topsoil.
 
5.~Respect wildlife: Do not interfere with the lives of birds and wild animals or disturb their peace.
\end{translationbox}

 
\subsection*{Example 7: Multi-Turn Conversation --- Mawlana Jalaluddin Balkhi (Rumi)}
\label{app:sample-multiturn-mawlana}
 
\begin{figure}[H]
    \centering
    \fcolorbox{gray!40}{gray!8}{\includegraphics[width=0.9\textwidth]{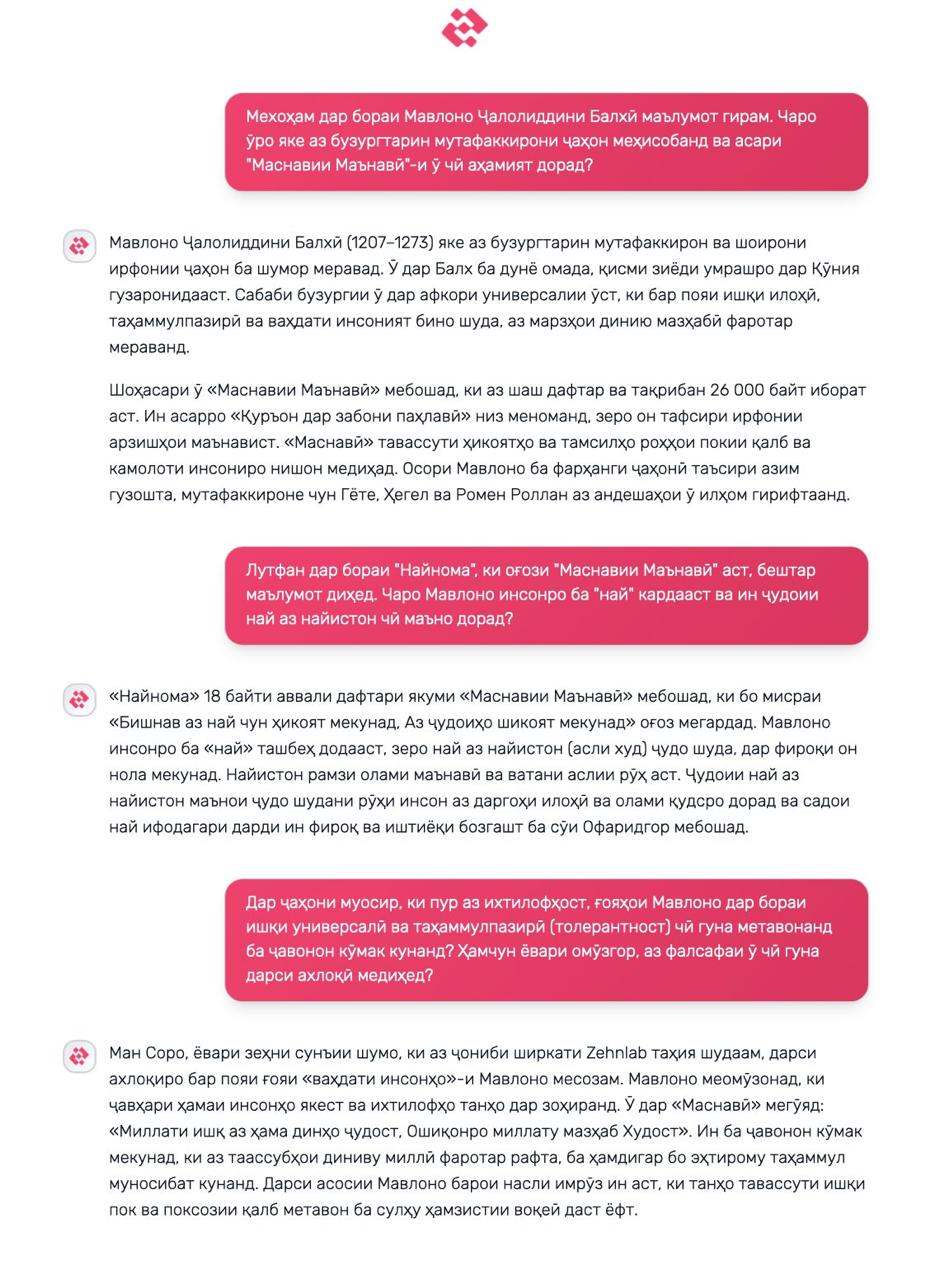}}
\end{figure}
 
\begin{translationbox}
\textbf{User (Turn 1):} I would like to receive information about Mawlana Jalaluddin Balkhi (Rumi). Why is he considered one of the world's greatest thinkers, and what is the significance of his work ``Masnavi-i Ma'navi''?
 
\medskip
\textbf{Soro (Turn 1):} Mawlana Jalaluddin Balkhi (1207--1273) is considered one of the greatest mystical thinkers and poets in the world. Born in Balkh, he spent much of his life in Konya. His greatness lies in his universal thoughts, built on divine love, tolerance, and the unity of humanity, which transcend religious and sectarian boundaries.
 
His masterpiece is ``Masnavi-i Ma'navi,'' which consists of six books and approximately 26,000 distichs (couplets). This work is also referred to as the ``Quran in the Pahlavi language'' because it is a mystical interpretation of spiritual values. The ``Masnavi'' uses stories and allegories to illustrate the paths of heart purification and human perfection. Mawlana's works have had a massive influence on world culture; thinkers such as Goethe, Hegel, and Romain Rolland were inspired by his ideas.
 
\medskip
\textbf{User (Turn 2):} Please tell me more about ``Naynoma,'' which begins the ``Masnavi-i Ma'navi.'' Why did Mawlana compare a human to a ``nay'' (reed flute), and what does this separation of the reed from the reedbed mean?
 
\medskip
\textbf{Soro (Turn 2):} ``Naynoma'' consists of the first 18 couplets of the first book of ``Masnavi-i Ma'navi,'' beginning with the line: ``Listen to the reed as it tells a tale, complaining of separations.'' Mawlana compares a human to a ``nay'' because the reed has been separated from the reedbed (its origin) and moans in longing for it. The reedbed symbolizes the spiritual world and the original home of the soul. The separation of the reed from the reedbed represents the separation of the human soul from the divine presence and the holy world; the sound of the reed expresses the pain of this separation and the yearning to return to the Creator.
 
\medskip
\textbf{User (Turn 3):} In the modern world, full of conflicts, how can Mawlana's ideas about universal love and tolerance help the youth? As a teacher's assistant, what ethical lesson would you give based on his philosophy?
 
\medskip
\textbf{Soro (Turn 3):} I am Soro, your AI assistant developed by Zehnlab, and I construct an ethical lesson based on Mawlana's idea of the ``unity of humanity.'' Mawlana teaches that the essence of all humans is one, and differences are only external. In the ``Masnavi,'' he says: ``The nation of love is separate from all religions; for lovers, the nation and religion is God.'' This helps young people look beyond religious and national prejudices and treat each other with respect and tolerance. Mawlana's main lesson for today's generation is that only through pure love and purification of the heart can true peace and coexistence be achieved.
\end{translationbox}


\section{Teacher Feedback Survey}
\label{app:teacher-survey}

To complement the qualitative observations reported in Section~\ref{sec:deployment}, we conducted a structured feedback survey among teachers participating in the Soro pilot deployment. This section presents the survey design, quantitative results, and a thematic summary of open-ended responses.

\subsection{Survey Design and Administration}

The survey was administered to teachers across a subset of 10 pilot schools during March 16--30, 2026, yielding 53 valid responses. Participating schools were located in Dushanbe, Vahdat, Khujand, Kulob, and Khorog, providing geographic coverage across all major deployment regions. Respondents taught a range of subjects, including primary education, English, Russian, informatics, IT, mathematics, biology, chemistry, history, and the newly introduced AI foundations course.

The survey comprised six Likert-scale items (rated 1--5, where 1~=~lowest and 5~=~highest) targeting key dimensions of the Soro deployment experience:

\begin{enumerate}[leftmargin=2em]
    \item \textbf{Response accuracy}: How accurately does Soro answer your questions?
    \item \textbf{Tajik language quality}: How well does Soro understand and use the Tajik language?
    \item \textbf{Lesson preparation usefulness}: How useful is Soro for lesson preparation (plans, materials, tests)?
    \item \textbf{Ease of use}: How easy is Soro to use?
    \item \textbf{Willingness to adopt}: How willing are you to use Soro in your daily work as a teacher?
    \item \textbf{Recommendation likelihood}: Would you recommend Soro to your colleagues?
\end{enumerate}

\noindent An optional seventh item solicited free-text comments and suggestions.

\subsection{Quantitative Results}

Table~\ref{tab:survey-results} summarizes the descriptive statistics for all six survey items. Across all questions, the overall mean rating is 4.27 out of 5 ($\sigma = 0.99$), and 79.9\% of all individual ratings are favorable (4 or 5). The highest-rated dimension is Tajik language quality (mean = 4.47, 86.8\% favorable), suggesting that the continual pretraining and instruction tuning pipeline described in Sections~\ref{sec:cpt}--\ref{sec:sft} is effective at producing fluent, natural Tajik output as perceived by native-speaking educators. Recommendation likelihood also scores highly (mean = 4.42, 84.9\% favorable), indicating broad willingness to endorse the tool among peers.

The lowest-rated dimensions are response accuracy (mean = 4.15) and willingness to adopt daily (mean = 4.11). While still favorable overall, these results corroborate the qualitative finding from Section~\ref{sec:deployment} that occasional factual errors on specific local topics reduce teacher confidence in unsupervised use. The relatively higher variance on willingness to adopt ($\sigma = 1.15$) reflects a polarization between enthusiastic early adopters and more cautious teachers who prefer to wait for further refinements.

\begin{table}[H]
\centering
\caption{Descriptive statistics for the teacher feedback survey ($n = 53$). Scores range from 1 (lowest) to 5 (highest). ``Favorable'' denotes the percentage of responses rated 4 or 5.}
\label{tab:survey-results}
\scalebox{0.85}{
\begin{tabular}{lcccc}
\toprule
\textbf{Survey Item} & \textbf{Mean} & \textbf{Std Dev} & \textbf{Median} & \textbf{Favorable (\%)} \\
\midrule
Response accuracy       & 4.15 & 0.99 & 4.0 & 75.5 \\
Tajik language quality  & 4.47 & 0.82 & 5.0 & 86.8 \\
Lesson prep usefulness  & 4.15 & 1.06 & 5.0 & 77.4 \\
Ease of use             & 4.34 & 1.02 & 5.0 & 81.1 \\
Willingness to adopt    & 4.11 & 1.15 & 5.0 & 73.6 \\
Recommend to colleagues & 4.42 & 0.84 & 5.0 & 84.9 \\
\midrule
\textbf{Overall}        & \textbf{4.27} & \textbf{0.99} & \textbf{5.0} & \textbf{79.9} \\
\bottomrule
\end{tabular}}
\end{table}

\begin{figure}[H]
    \centering
    \begin{tikzpicture}
    \begin{axis}[
        xbar,
        /pgf/bar shift=0pt,
        width=0.82\textwidth,
        height=6.5cm,
        xlabel={Mean Rating (1--5)},
        symbolic y coords={
            {Willingness to adopt},
            {Response accuracy},
            {Lesson prep usefulness},
            {Ease of use},
            {Recommend to colleagues},
            {Tajik language quality}
        },
        ytick=data,
        yticklabel style={font=\small},
        nodes near coords,
        nodes near coords align={horizontal},
        nodes near coords style={font=\footnotesize},
        every node near coord/.append style={
            /pgf/number format/.cd,
            fixed,
            precision=2,
            /tikz/.cd
        },
        xmin=1,
        xmax=5.5,
        bar width=0.5cm,
        enlarge y limits=0.12,
        axis line style={-},
        tick style={draw=none},
        ymajorgrids=false,
        xmajorgrids=true,
        grid style={dashed, gray!30},
        error bars/x dir=both,
        error bars/x explicit,
    ]
    
    \addplot[fill=soromain, draw=soromainborder,
        error bars/error bar style={thick, soromainborder},
        error bars/error mark options={rotate=90, mark size=3pt, thick, soromainborder}
    ] coordinates {
        (4.11,{Willingness to adopt})     +- (1.15,0)
        (4.15,{Response accuracy})        +- (0.99,0)
        (4.15,{Lesson prep usefulness})   +- (1.06,0)
        (4.34,{Ease of use})              +- (1.02,0)
        (4.42,{Recommend to colleagues})  +- (0.84,0)
        (4.47,{Tajik language quality})   +- (0.82,0)
    };
    
    \end{axis}
    \end{tikzpicture}
    \caption{Mean teacher ratings across six survey dimensions ($n = 53$). Error bars indicate one standard deviation.}
    \label{fig:survey-mean-ratings}
\end{figure}
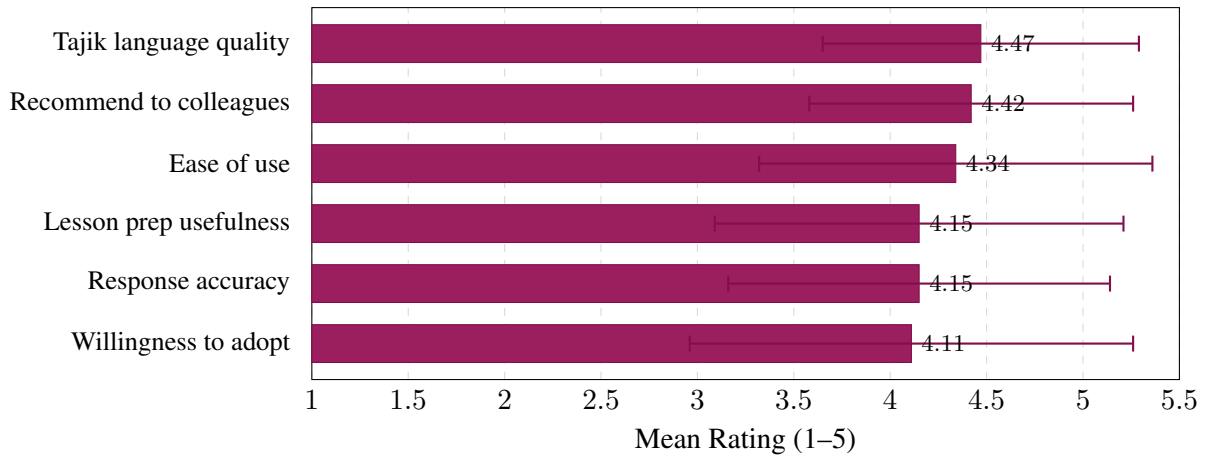

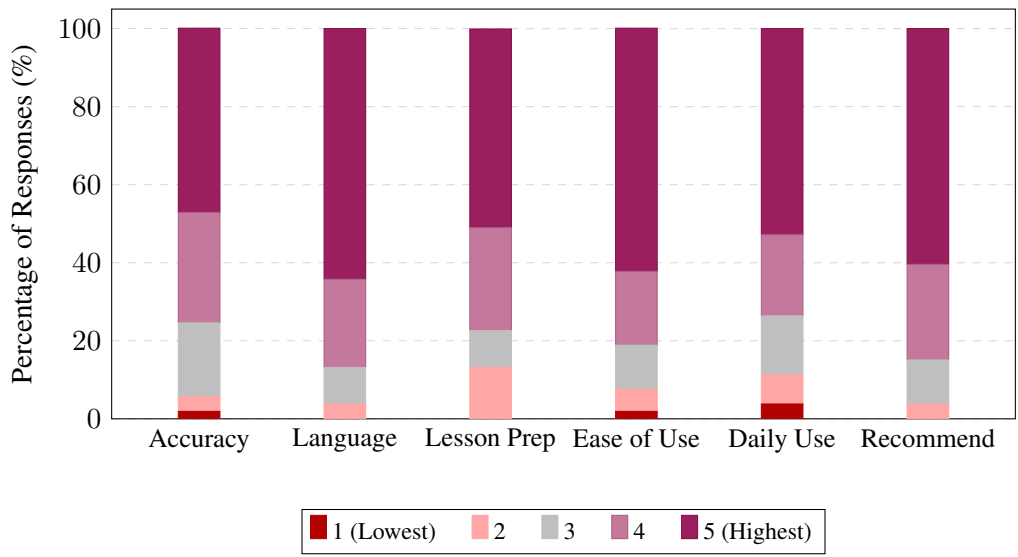
\begin{figure}[H]
    \centering
    \begin{tikzpicture}
    \begin{axis}[
        ybar stacked,
        width=0.85\textwidth,
        height=7cm,
        ylabel={Percentage of Responses (\%)},
        symbolic x coords={
            {Accuracy},
            {Language},
            {Lesson Prep},
            {Ease of Use},
            {Daily Use},
            {Recommend}
        },
        xtick=data,
        xticklabel style={font=\small},
        ymin=0,
        ymax=105,
        bar width=0.55cm,
        enlarge x limits=0.12,
        axis line style={-},
        tick style={draw=none},
        ymajorgrids=true,
        grid style={dashed, gray!30},
        legend style={
            at={(0.5,-0.22)},
            anchor=north,
            legend columns=5,
            font=\footnotesize,
            /tikz/every even column/.append style={column sep=0.4cm},
        },
        legend cell align={left},
    ]
    
    \addplot[fill=red!70!black, draw=red!70!black] coordinates {
        ({Accuracy},     1.9)
        ({Language},     0.0)
        ({Lesson Prep},  0.0)
        ({Ease of Use},  1.9)
        ({Daily Use},    3.8)
        ({Recommend},    0.0)
    };
    
    \addplot[fill=red!35!white, draw=red!35!white] coordinates {
        ({Accuracy},     3.8)
        ({Language},     3.8)
        ({Lesson Prep}, 13.2)
        ({Ease of Use},  5.7)
        ({Daily Use},    7.5)
        ({Recommend},    3.8)
    };
    
    \addplot[fill=gray!50, draw=gray!50] coordinates {
        ({Accuracy},    18.9)
        ({Language},     9.4)
        ({Lesson Prep},  9.4)
        ({Ease of Use}, 11.3)
        ({Daily Use},   15.1)
        ({Recommend},   11.3)
    };
    
    \addplot[fill=sorofamily, draw=sorofamilyborder] coordinates {
        ({Accuracy},    28.3)
        ({Language},    22.6)
        ({Lesson Prep}, 26.4)
        ({Ease of Use}, 18.9)
        ({Daily Use},   20.8)
        ({Recommend},   24.5)
    };
    
    \addplot[fill=soromain, draw=soromainborder] coordinates {
        ({Accuracy},    47.2)
        ({Language},    64.2)
        ({Lesson Prep}, 50.9)
        ({Ease of Use}, 62.3)
        ({Daily Use},   52.8)
        ({Recommend},   60.4)
    };
    
    \legend{1 (Lowest), 2, 3, 4, 5 (Highest)}
    
    \end{axis}
    \end{tikzpicture}
    \caption{Distribution of teacher ratings across survey dimensions ($n = 53$). Each bar shows the percentage breakdown by score.}
    \label{fig:survey-distribution}
\end{figure}

Figure~\ref{fig:survey-mean-ratings} presents the mean ratings with standard deviation error bars across all six dimensions, and Figure~\ref{fig:survey-distribution} shows the full rating distribution for each item.

\subsection{Qualitative Feedback}

Of the 53 respondents, 25 provided optional free-text comments. A thematic analysis of these responses reveals three recurring categories.

\paragraph{Positive reception.} 
A majority of commenters expressed enthusiasm about the availability of a Tajik-language AI assistant. Representative remarks highlight Soro as a valuable aid for teachers, with several noting hopes that it will reduce the burden of manual lesson preparation.

\paragraph{Requests for improvement.} 
The most frequently cited concern was response latency at the interaction level: multiple teachers reported that Soro sometimes takes too long to generate answers for live classroom use, even though the centralized service itself remained generally available during the pilot. A second recurring theme was response completeness; some teachers found answers too brief or lacking in specificity, particularly on domain-specific queries. One respondent noted that responses occasionally rely on outdated information and lack follow-up suggestions. These observations are consistent with the quantitative finding that response accuracy received the lowest favorable percentage (75.5\%) and are actively informing ongoing improvements to the model and inference infrastructure.

\paragraph{Feature requests.} 
Teachers requested voice interaction capabilities, more detailed responses on specialized topics, subject-specific test generation with answer keys, and curriculum-aligned lesson plan templates approved by the Ministry of Education and Science of Tajikistan.

\subsection{Discussion}

The survey results provide preliminary evidence that Soro is perceived positively by frontline educators. The strong rating on Tajik language quality (4.47/5) is consistent with the core claim of this work: that targeted continual pretraining on a curated Tajik corpus, followed by instruction tuning, produces output that native speakers find natural and fluent. The high recommendation likelihood (4.42/5) suggests organic adoption potential beyond the current pilot scope.

The feedback also highlights concrete areas for improvement. Latency concerns motivate further optimization of inference infrastructure, including more aggressive quantization and batching strategies for the deployed 27B model. Requests for deeper, more specific responses suggest that expanding the instruction tuning dataset with additional subject-specific examples, particularly in STEM and social sciences, could improve perceived utility. The interest in voice interaction aligns with planned development of multimodal capabilities for future Soro iterations.

Several limitations of this survey deserve acknowledgment. The sample size ($n = 53$) is modest and drawn from a convenience sample of early-adopting pilot schools, which may introduce positive selection bias. Respondents were not randomly selected, and the survey was administered during the initial deployment phase when novelty effects may inflate ratings. A more rigorous, longitudinal evaluation with control groups and objective learning outcome measures is planned for the next phase of the pilot expansion.

\section{Tajikistan National Curriculum: Subjects in Grades 5--11}
\label{app:curriculum}

This section provides an overview of the subjects taught in Tajikistan's general secondary schools across grades 5--11. This curricular structure contextualizes both the educational data subset used in continual pretraining (Section~\ref{subsec:dataset-for-continual-pretraining}) and the deployment use cases described in Section~\ref{sec:deployment}, as Soro is designed to serve as a learning assistant aligned with the national educational framework.

\subsection{Structure of Secondary Education in Tajikistan}

General secondary education in Tajikistan is organized into three stages: primary school (grades 1--4), basic secondary school (grades 5--9), and upper secondary school (grades 10--11). Basic secondary education (grades 1--9) is compulsory and free of charge. Upon completion of grade 9, students receive a certificate of basic general education. After completing basic secondary education, learners are required to continue their studies either in general secondary school (grades 10--11) or in primary or secondary vocational education; admission to grades 10--11 is based on learner preferences and academic performance.

Upper secondary education culminates in the Centralized University Entrance Examinations (\textit{Markazi millii testī}), administered by the National Testing Center under the President of the Republic of Tajikistan. These examinations serve as the primary gateway to higher education and assess knowledge across multiple subjects from the national curriculum.

The curriculum follows a competency-based education (CBE) framework introduced in 2015--2017, which transitioned the system from a knowledge-transmission model toward a more student-centered, competency-oriented approach. Despite this reform, the subject structure retains continuity with the Soviet-inherited disciplinary tradition common across Central Asian education systems, with clearly delineated courses in languages, mathematics, natural sciences, social sciences, and arts.

\subsection{Subject Areas}

The following lists the core subjects prescribed by the national curriculum for grades 5--11. Subject names are provided in English with Tajik transliterations. Some subjects span the full grades 5--11 range, while others are introduced at specific stages (e.g., Physics from grade 7) or undergo structural changes (e.g., unified Mathematics splitting into Algebra and Geometry).

\paragraph{Languages and Literature.}
\begin{itemize}
    \item \textbf{Tajik Language} (\textit{Zaboni tojiki}) - grammar, morphology, syntax, and orthography of the state language; taught throughout grades 5--11.
    \item \textbf{Tajik Literature} (\textit{Adabiyot}) - classical and contemporary Tajik-Persian literary tradition, including poetry, prose, and literary analysis; grades 5--11.
    \item \textbf{Russian Language} (\textit{Zaboni rusi}) - mandatory second language; grades 5--11.
    \item \textbf{Foreign Language} (\textit{Zaboni khoriji}) - primarily English, in some schools German or French; grades 5--11.
\end{itemize}

\paragraph{Mathematics and Computer Science.}
\begin{itemize}
    \item \textbf{Mathematics} (\textit{Matematika}) - unified course covering arithmetic, pre-algebra, and introductory geometry; grades 5--6.
    \item \textbf{Algebra} (\textit{Algebra}) and \textbf{Geometry} (\textit{Geometriya}) - separate courses from grade 7 through grade 11, progressively covering functions, equations, trigonometry, spatial geometry, and introductory calculus.
    \item \textbf{Informatics / ICT} (\textit{Informatika}) - fundamentals of computing, digital literacy, and basic programming; introduced in basic secondary school and continuing through grade 11.
\end{itemize}

\paragraph{Natural Sciences.}
\begin{itemize}
    \item \textbf{Biology} (\textit{Biologiya}) - botany, zoology, human anatomy, and general biology; grades 5/6--11.
    \item \textbf{Physics} (\textit{Fizika}) - mechanics, thermodynamics, electricity, optics, and nuclear physics; introduced from grade 7.
    \item \textbf{Chemistry} (\textit{Khimiya}) - inorganic and organic chemistry; introduced from grade 7 or 8 depending on the school program.
    \item \textbf{Geography} (\textit{Jughrofiya}) - physical and economic geography, including a dedicated Geography of Tajikistan component; grades 5/6--11.
\end{itemize}

\paragraph{Social Sciences and Humanities.}
\begin{itemize}
    \item \textbf{History of the Tajik People} (\textit{Ta'rikhi khalqi tojik}) - national history from ancient Central Asian civilizations through the Samanid dynasty, Soviet era, and post-independence period; grades 5--9, followed by broader history courses in grades 10--11.
    \item \textbf{Social Studies / Civics} (\textit{Jam'iyatshinosi}) - foundations of society, government, and civic responsibility; taught in upper grades of basic and upper secondary school.
    \item \textbf{Legal Studies / Basics of State and Law} (\textit{Huquqshinosi / Asoshoi davlat va huquq}) - foundations of law, rights and responsibilities, and basic civic-legal concepts; typically taught in upper grades.
\end{itemize}

\paragraph{Arts, Technology, and Physical Education.}
\begin{itemize}
    \item \textbf{Music} (\textit{Musiqi}) and \textbf{Fine Arts} (\textit{San'ati tasviri}) - primarily taught in grades 5--7, covering Tajik musical heritage and visual arts.
    \item \textbf{Labor / Technology} (\textit{Mehnat / Tekhnologiya}) - practical skills and vocational orientation; grades 5--9.
    \item \textbf{Physical Education} (\textit{Tarbiyai jismoni}) - throughout grades 5--11.
    \item \textbf{Initial Military Training} (\textit{Omuzishi ibtidoii harbi}) - conducted with male students during the last two years of schooling (grades 10--11) according to the approved curriculum.
\end{itemize}

\subsection{Relevance to the Soro Training Corpus and Deployment}

The educational resources subset of our pretraining corpus (Section~\ref{subsec:dataset-for-continual-pretraining}) consists of scanned and manually transcribed textbooks from the subjects listed above, spanning grades 5--11, sourced exclusively from officially approved Tajik school materials to ensure alignment with national curriculum content, terminology, and pedagogical framing.

The breadth of subject coverage is deliberate. Humanities subjects, particularly History of the Tajik People, Tajik Literature, and Geography of Tajikistan, provide culturally grounded knowledge, domain-specific vocabulary, and historical narratives that are largely absent from general-purpose multilingual training corpora; this content directly supports the Tajik-FactQA and Tajik History benchmarks (Appendix~\ref{app:tajik-history}). STEM subjects (Physics, Chemistry, Biology, Mathematics) complement the translated FineWeb-Edu subset and ensure the model acquires scientific and technical terminology in Tajik. The linguistic subjects (Tajik Language, Tajik Literature) reinforce grammatical and lexical competence, contributing to the strong gains on the TajLib benchmark (Appendix~\ref{app:tajlib}).

From a deployment perspective (Section~\ref{sec:deployment}), alignment between the training corpus and the national curriculum is practical rather than merely principled: students and teachers interacting with Soro expect it to be knowledgeable about the subjects they study and teach. Curriculum-aligned training ensures that responses are not only linguistically fluent but also factually consistent with the materials used in Tajik classrooms.

\end{document}